\documentclass{article}

\usepackage[final]{corl_2020} 

\usepackage{times}
\usepackage[utf8]{inputenc} 
\usepackage[T1]{fontenc}    
\usepackage{url}            
\usepackage{booktabs}       
\usepackage{nicefrac}       
\usepackage{microtype}      
\usepackage{graphics,color,subcaption}

\usepackage[ruled,vlined,linesnumbered]{algorithm2e}
\DontPrintSemicolon
\SetAlgoLined
\usepackage{enumitem}

\usepackage{amsfonts}       
\usepackage{amsmath}       
\usepackage{amssymb}

\newcommand{\Fig}[1]{Figure~\ref{#1}}  
\newcommand{\fig}[1]{Fig.~\ref{#1}}    
\newcommand{\tab}[1]{Table~\ref{#1}}

\newcommand{\eqn}[1]{Eq.~\ref{#1}} 
\renewcommand{\sec}[1]{Sec.~\ref{#1}} 

\newcommand{\algo}[1]{Alg.~\ref{#1}}    

\usepackage{xspace}
\makeatletter
\DeclareRobustCommand\onedot{\futurelet\@let@token\@onedot}
\def\@onedot{\ifx\@let@token.\else.\null\fi\xspace}
\def\eg{e.g\onedot}
\def\ie{i.e\onedot}

\def\wrt{w.r.t\onedot}
\makeatother

\newcommand{\Unit}{\ensuremath{\mathbb I}}        
\newcommand\Tstrut{\rule{0pt}{2.4ex}}

\usepackage[dvipsnames]{xcolor}
\definecolor{ourblue}{rgb}{0.368,0.507,0.71}
\definecolor{ourorange}{rgb}{0.881,0.611,0.142}
\definecolor{ourgreen}{rgb}{0.56,0.692,0.195}
\definecolor{ourred}{rgb}{0.923,0.386,0.209}
\definecolor{ourviolet}{rgb}{0.528,0.471,0.701}
\definecolor{ourbrown}{rgb}{0.772,0.432,0.102}
\definecolor{ourlightblue}{rgb}{0.364,0.619,0.782}
\definecolor{ourdarkgreen}{rgb}{0.572,0.586,0.}

\graphicspath{{graphics/}}

\newcommand{\beginsupplement}{
	\renewcommand{\thefigure}{S\arabic{figure}}
	\renewcommand{\thetable}{S\arabic{table}}
}

\usepackage[disable]{todonotes}

\newcommand{\Cri}[2][inline]{\todo[#1,color=red!25,size=\scriptsize]{#2}}

\newcommand{\Method}{\textrm{iCEM}}
\newcommand{\CEMMPC}{$\text{CEM}_{\text{MPC}}$}
\newcommand{\CEMPETS}{$\text{CEM}_{\text{PETS}}$}
\newcommand{\env}[1]{\textsc{#1}}
\newcommand{\HS}{\env{Humanoid Standup}}
\newcommand{\HCR}{\env{Halfcheetah Running}}
\newcommand{\FPP}{\env{Fetch Pick\&Place}}
\newcommand{\Door}{\env{Door}}
\newcommand{\DoorNS}{\env{Door} (sparse reward)}
\newcommand{\Reloc}{\env{Relocate}}

\usepackage{bbm}
\usepackage{footmisc}

\usepackage{listings}
\usepackage{multirow}

\usepackage[export]{adjustbox}
\usepackage{tabularx}
\usepackage{collcell}

\newcommand\clearrow{\global\let\rowmac\relax}
\clearrow

\newcolumntype{C}{>{\collectcell\rowmac}c<{\endcollectcell}}
\newcolumntype{R}{>{\collectcell\rowmac}r<{\endcollectcell}}
\newcolumntype{L}{>{\collectcell\rowmac}l<{\endcollectcell}}

\newcommand{\rpm}{\raisebox{.2ex}{$\scriptstyle\pm$}}
\usepackage{authblk}
    
\usepackage{hyperref}

\hypersetup{hidelinks,
backref=true,
pagebackref=true,
hyperindex=true,
breaklinks=true,
colorlinks=true,linkcolor=ourblue,
urlcolor=ourblue,citecolor=ourviolet,
bookmarks=true,
bookmarksopen=false}
\usepackage{relsize}

\title{\LARGE \bf
Sample-efficient Cross-Entropy Method \\for Real-time Planning} 

\author[1]{Cristina Pinneri\thanks{Further affiliation: Max Planck ETH Center for Learning Systems, T\"ubingen - Z\"urich}\ \,}
\author[1]{Shambhuraj Sawant}
\author[1]{Sebastian Blaes}
\author[2]{Jan Achterhold}
\author[2]{J\"org St\"uckler}
\author[1]{Michal Rol\'inek}
\author[1]{Georg Martius}
\affil[1]{Autonomous Learning Group\\
    Max Planck Institute for Intelligent Systems\\
    Germany\\
    T\"ubingen
}
\affil[2]{Embodied Vision Group\\
    Max Planck Institute for Intelligent Systems\\
    Germany\\
    T\"ubingen\\    \texttt{firstname.lastname@tuebingen.mpg.de}
}

\makeatletter
    \renewcommand{\@noticestring}{}
\makeatother

\begin{document}

\maketitle
\thispagestyle{empty}
\pagestyle{empty}

\begin{abstract} 
Trajectory optimizers for model-based reinforcement learning, such as the  Cross-Entropy Method (CEM), can yield compelling results even in high-dimensional control tasks and sparse-reward environments.
However, their sampling inefficiency prevents them from being used for real-time planning and control.
We propose an improved version of the CEM algorithm for fast planning, with novel additions including temporally-correlated actions and memory,
requiring 2.7-22$\times$ less samples and yielding a performance increase of 1.2-10$\times$ in high-dimensional control problems.
\end{abstract}


\begin{figure}[h!]
    \centering
    \includegraphics[height=0.098\linewidth]{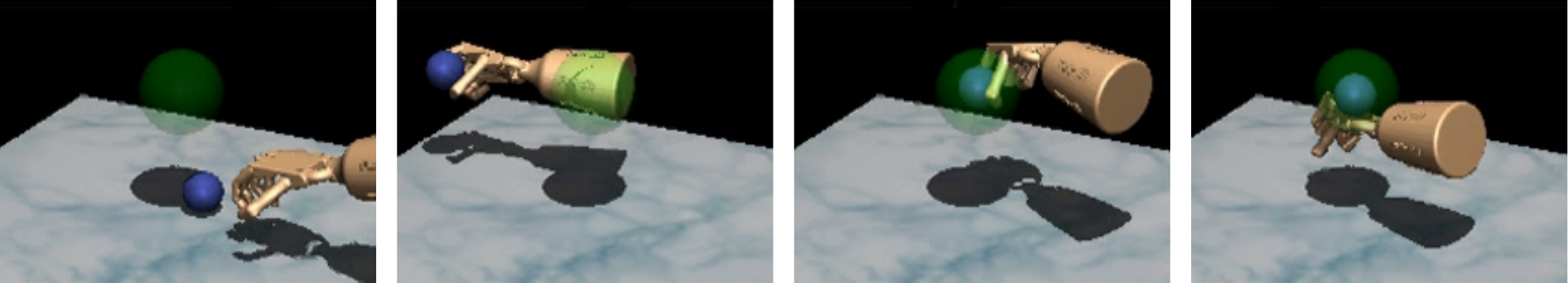}\quad
    \includegraphics[height=0.098\linewidth]{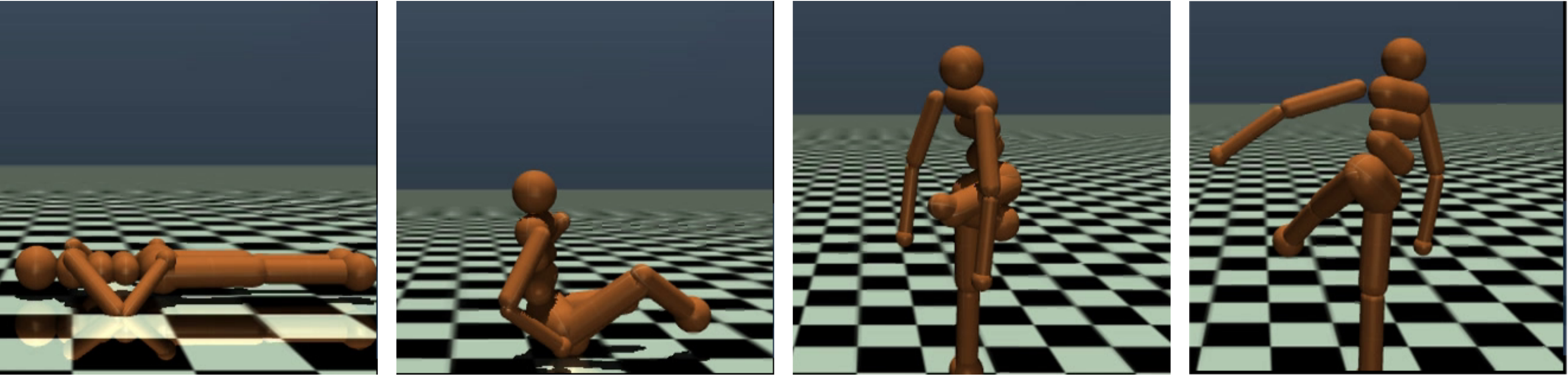}
    \caption{Found behaviors for \Reloc{} environment (left) and \HS{} (right)}
    \label{fig:environments}
\end{figure}

\section{Introduction}\label{sec:intro}

Recent work in model-based reinforcement learning (MBRL) for high-dimensional systems employs population-based algorithms as trajectory optimizers \cite{chua2018pets, ba2019poplin, hafner2018planet, Williams2015MPPI, naga2018mbmf}.
Sampling-based methods have also been used in the control community in scenarios when the cost function is not differentiable \cite{richards2006RS}.
The particular appeal of these methods lies in a few but important factors: the possibility of optimizing black-box functions; lower sensitivity to hyperparameter tuning and thus higher robustness; no requirement of gradient information; lower susceptibility to local optima.
The Cross-Entropy Method (CEM) \cite{Rubinstein99cem} was introduced for the first time in the 1990s as a stochastic, derivative-free, global optimization technique, but it is just in recent years that it gained traction in the model-based RL community.
CEM for trajectory optimization is indeed a promising metaheuristics which has been shown to work well even with learned models,
 producing comparable or higher performance than model-free reinforcement learning methods~\cite{chua2018pets, ba2019poplin, hafner2018planet}.

There is a problem, however, intrinsic to the nature of population-based optimizers, which makes these methods so far unsuitable for \textbf{real-time planning and control}, even in conjunction with a learned model: the high computational price.
Heuristics like CEM require a large number of samples to minimize the objective function.\Cri[]{CEM estimates the probability distribution over the solution space which concentrates on the global maxima of the objective function}
This creates severe limitations for its deployment in real-time control for robotics, requiring a dramatic speed-up.

Our approach originates exactly from this question: is it possible to do real-time planning with a zeroth-order optimizer like CEM?
Our method proposes an enhancement of the original CEM for the purpose of trajectory optimization in model-predictive control and comprises various ways to address the inefficiency of sampling in high-dimensional systems, including equipping CEM with memory and generating time-correlated action sequences. Our upgrades are unified under the name \textbf{\Method{}}.

\textbf{Contributions} \quad
We present \Method{}, a faster, more sample-efficient and higher performing version of the CEM algorithm that could potentially bridge the gap between MBRL in simulation and real-time robotics.
We present a detailed examination of the key improvements over CEM with an extensive ablation study.
Finally, we test the results on several hard continuous-control robotic tasks in the MuJoCo simulator \cite{todorov2012mujoco} such as \HS{}, and manipulation environments with sparse rewards like \FPP{} or
 other manipulation environments with many degrees of freedom like \Door{} and \Reloc{}. In the latter, we solve the task with 90\% success rate while using 13.7$\times$ less samples and get an average performance improvement of 400\% over the state-of-the-art CEM.

In order to study the algorithmic improvements without being biased by model errors, we perform all our ablations with the ground truth dynamics.
In addition to this, we report the performance when used in combination with learned models from a reimplementation of the PlaNet framework \cite{hafner2018planet} (without requiring additional fine-tuning), showing a speed-up that potentially allows online planning with \Method{}, without a substantial loss on the overall performance.

To the best of our knowledge, this is the first work that aims at making CEM itself fast enough to be used for real-time robot planning and control. It can be integrated into any existing method that uses the standard CEM or other zeroth-order optimizers.
The source code will be made available with the final version of the paper.

\textbf{Related Work}\quad
Many works on MBRL and motion planning show that it is possible to control systems without making use of gradient descent.
Indeed, Evolution Strategies (ES) \cite{rechenberg1971ES} regained popularity for their successful use in RL \cite{salimans2017evolution, choromanski2018policyES, mania2018simple}, making population-based methods an attractive alternative to policy gradient or as a supportive guidance \cite{khadka2018evolutionguided, bharadhwaj2020modelpredictive}.
Sampling-based techniques have also been used for model-predictive control (MPC), like model predictive path integral (MPPI) control \cite{Williams2015MPPI}, with applications to aggressive driving by using a GPU \cite{Williams2016AggressiveDW}.

In particular, the Cross-Entropy Method (CEM) \cite{Rubinstein99cem, rubinstein2004cem, botev2013cem}, thoroughly analyzed in \cite{margolin2005CEManalysis}, has been used both for direct policy optimization \cite{pourchot2018cemrl} and planning with learned models \cite{chua2018pets, ba2019poplin}, to improve the performance of rapidly exploring random trees \cite{kobilarov2012motionplanning}, with successful applications in many fields of science. Examples include visual tracking \cite{cehovin2011cemvisualtracking}, bioinformatics \cite{lin2008cembioinfo}, and network reliability \cite{Hui2005cemnetwork}.

In \citet{duan2016benchmarking} it is reported that CEM has better performance also over the more sophisticated Covariance Matrix Adaptation ES (CMA-ES) \cite{hansen1996CMAES}, the latter being computationally more expensive since it computes the full covariance matrix, while actions in CEM are sampled independently along the planning horizon, requiring only a diagonal covariance matrix. \Cri[]{try to find more modifications to CEM to add}

The core focus of our work is to make CEM functional for real-time decision making. Recent works in this direction propose a differentiable version of CEM \cite{br2019differentiable} or to jointly use model gradients together with the CEM search \cite{bharadhwaj2020modelpredictive}.
\citet{ba2019poplin} use CEM on the policy parameters rather than in the action space. Nevertheless, the whole procedure still depends on the speed of the CEM optimization, making it the bottleneck for fast planning.

\section{The Cross-Entropy Method}\label{sec:CEM}
The cross-entropy method (CEM) is a derivative-free optimization technique that was originally introduced in \citet{Rubinstein99cem} as an adaptive importance sampling procedure for the estimation of rare-event probabilities that makes use of the \textit{cross-entropy} measure.

CEM can be seen as an Evolution Strategy which minimizes a cost function $f(x)$ with $f: \mathbb{R}^n \to \mathbb{R}$ by finding a suitable ``individual'' $x$.

The individuals are sampled from a population/distribution and evaluated according to $f(x)$.
 Then, they are sorted based on this cost function and a fixed number of ``elite'' candidates is selected.

 This \emph{elite-set} is going to determine the parameters of the population for the next iteration.
 In the standard case, the population is modeled with a Gaussian distribution with mean $\mu$ and diagonal covariance matrix $\textrm{diag}(\sigma^2)$, where $\mu, \sigma \in \mathbb{R}^n$.
By fitting $\mu$ and $\sigma$ to the elite-set, the sampling distribution concentrates around the $x$ with low cost.
After several iterations of this selection procedure, an $x$ close to a local optimum, or even the global optimum, is found.
Due to this iterative procedure, the total number of evaluated samples becomes extensive which can lead to a slow run time depending on the computational cost of $f$.

\subsection{Standard modifications of CEM for model-predictive control: $\text{CEM}_{\text{MPC}}$}

In the MPC setting, CEM is used every timestep to optimize an $h$-step planning problem on the action sequences,
see \algo{algo:cem}.
For the terminology, we call \emph{CEM-iteration} one step of the inner loop of CEM that optimizes the sampling distribution (line \ref{algo:cem:inner}--\ref{algo:cem:inner:end} in \algo{algo:cem}).
The outer loop \emph{step} marks the progression in the environment by executing one action.
Naturally, the next step considers the planning problem one timestep later.
As a typical modification \cite{chua2018pets,ba2019poplin}, the initial mean $\mu_t$ of the CEM distribution is shift-initialized for the next time step
 using the one optimized in the previous time step $\mu_{t-1}$. Both $\mu_t, \sigma_t \in \mathbb{R}^{d \times h}$, with $d$ dimensionality of the action space.

Another standard modification is to use a momentum term \cite{de2005tutorial} in the refitting of the distributions between the CEM-iterations (line \ref{algo:cem:refit} in \algo{algo:cem}).
The reason is that only a small elite-set is used to estimate many parameters of the sampling distribution. A simple choice is
$\mu_t^{i+1} = \alpha \mu_{t}^i + (1-\alpha)\mu^{elite-set_i}$ where $\alpha \in [0,1]$ and $i$ is the index of \emph{CEM-iterations}.
Actions are always limited such that the standard method uses truncated normal distributions with suitably adapted bounds instead of unbounded Gaussian distributions.
We call this variance {\bf \CEMMPC}.

In \citet{chua2018pets} (PETS method), in addition to the standard improvements above, the sampling distribution was modified (only documented in the source code).
Instead of setting the truncation bounds to match the action-range, the truncation is always set to $2\sigma$,
and $\sigma$ is adapted to be not larger than $\frac{1}{2} b$ where $b$ is the minimum distance to the action bounds.
We refer to this method as {\bf \CEMPETS}.

\section{Improved CEM -- \Method}\label{sec:methods} 
In this section we thoroughly discuss several improvements to CEM for the purpose of model-predictive control (MPC) and trajectory optimization,
with the goal to achieve strong performance already with a low number of samples.
This section is complemented by the ablations in \sec{sec:ablations}
and the sensitivity analysis in \sec{sec:sup:sensitivity}.

\subsection{Colored noise and correlations}\label{subsec:colored-noise}
The CEM action samples should ideally produce trajectories which maximally explore the state space, especially if the rewards are sparse.
Let us consider a simple stochastic differential equation \Cri[]{double-check this} in which the trajectory $x$ is a direct integration of the stochastic actions $a$:
\begin{equation}\label{eq:RW}
    \frac{\textrm{d}}{\textrm{d}t}x(t)=a(t)
\end{equation}
In the case of Gaussian inputs, $x(t)$ is a Brownian random walk, which is commonly used to describe the trajectory of particles under random perturbations.
It comes as no surprise that coherent trajectories cannot be generated by uncorrelated inputs, like the ones sampled in CEM.

It was witnessed many times in nature that animals revert to different strategies, rather than plain Brownian exploration, when they need to efficiently explore the space in search for food. In fact, when prey is scarce, animals like sharks or other predatory species produce trajectories which can be described by so called L\'{e}vy walks \cite{humphries2010levysharks}.
Classically, L\'{e}vy walks exhibit velocities with long-term correlations (being sampled from a power-law distribution), and consequentially produce trajectories with higher variance than a Brownian motion.

If we look at the action sequence as a time series, its correlation structure is directly connected to the power spectral density (PSD) as detailed in \sec{sec:sup:colored-noise} in the Supplementary. The PSD is the squared norm of the value of Fourier transform and intuitively quantifies how much each frequency is occurring in the time series.
The CEM actions, being sampled independently along the planning horizon, have a constant power spectral density (PSD), more commonly referred to as \emph{white-noise}.
How does the PSD of a time series with non-zero correlations look like?
For this purpose, we introduce generalized colored-noise for the actions $a$ as the following PSD:
\begin{equation}\label{eq:PSD-noise}
    \textrm{PSD}_a(f) \propto \frac{1}{f^\beta}
\end{equation}
where $f$ is the frequency and $\beta$ is the colored-noise scaling exponent. $\beta=0$ corresponds to white noise, a value of $\beta>0$ means that high frequencies are less prominent than low ones.
In signal processing they are called \emph{colored noise} with pink noise for $\beta=1$, and Brownian or red noise for $\beta=2$, but any other exponent is possible. \goodbreak
\begin{figure}
\centering
  \begin{tabular}{c@{\ }c}
    (a) 1D random walks with colored noise & (b) power spectral density of actions\\
    \includegraphics[height=0.26\linewidth]{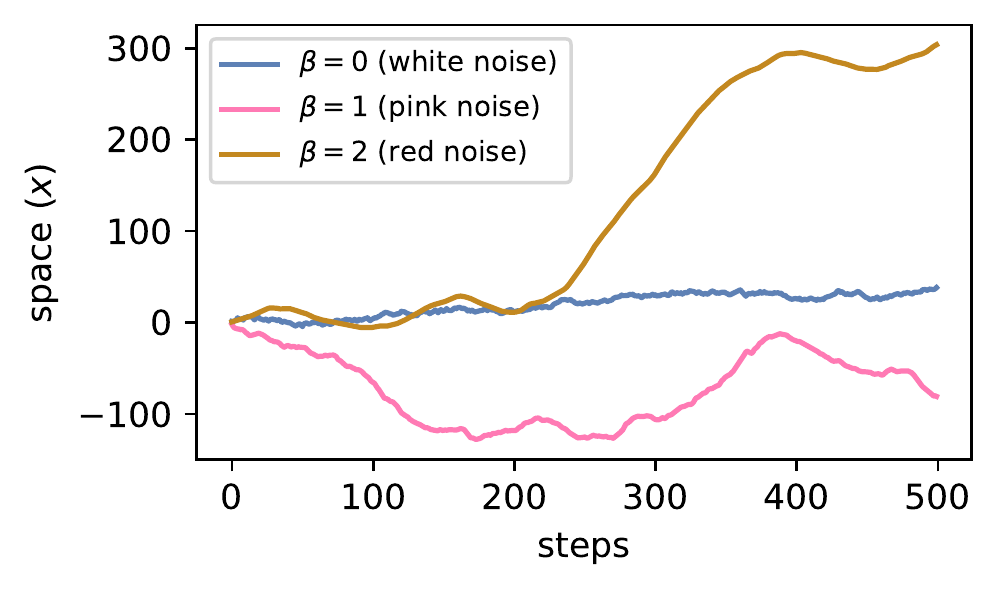}&
    \includegraphics[height=0.26\linewidth]{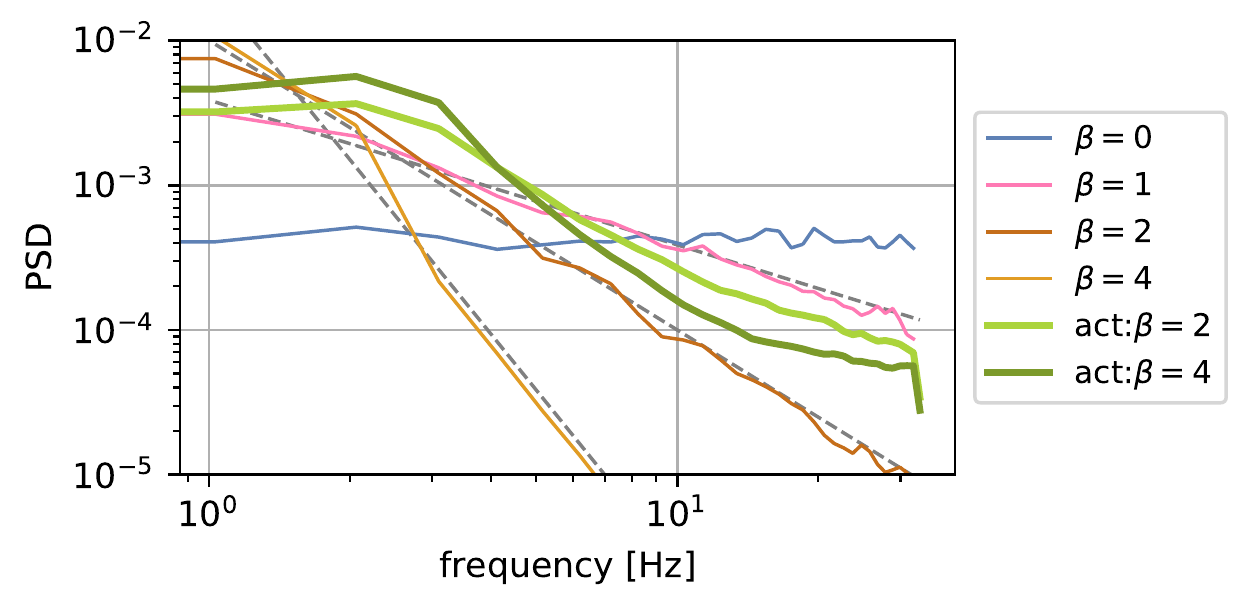}
  \end{tabular}
  \caption{Colored random noise. (a) random walks with colored noise of different temporal structure. (b) power spectrum of colored-random action sequences for different $\beta$ and of the chosen (and successful) action-sequences (labeled with \emph{act}) of \Method{} with differently colored search noise for the \HS{} task.
  Successful action sequences are far from white-noise ($\beta=0$).}
\label{fig:noise-psd}
\end{figure}

How does the trajectory $x(t)$ of \eqn{eq:RW} look when we use colored-noise actions? \Fig{fig:noise-psd}(a) shows three examples with the same action variance -- the larger the $\beta$, the larger the coherence and the larger distances can be reached.
This can be formalized by computing the $\textrm{PSD}_x$ of the state-space trajectory ($x$). Using \eqn{eq:RW} and \eqn{eq:PSD-noise} we find:
\begin{equation}\label{eq:PSD_RW}
\textrm{PSD}_x(f) = \|\mathcal{F}[x(t)](f)\|^2 \stackrel{(*)}{=} \frac{\|\mathcal{F}[a(t)](f)\|^2}{4 \pi^2 f^2} = \frac{\textrm{PSD}_{a}(f)}{4 \pi^2 f^2} \propto \frac{1}{f^{\beta + 2}}.
\end{equation}
where the equality $(*)$ results from the integration property of Fourier transforms, which is $\mathcal{F}[\frac{\textrm{d}}{\textrm{d}t}x(t)]=i 2\pi f \mathcal{F}[x(t)]$.
As a result, the PSD of $x(t)$ is directly controlled by the choice of $\beta$ -- higher $\beta$ results in stronger low frequency components, as evident in \fig{fig:noise-psd}(a).

Let us consider now the effect of colored-noise in a robotic setting:
 the \HS{} task, see \fig{fig:environments}, included in the OpenAI Gym~\cite{brockman2016openai} environments.
\Fig{fig:noise-psd}(b) displays the PSD$_a$ of different action-noise processes together with the PSD of successful action-sequences.
Notice the log-log scale -- a straight line corresponds to a power-law decay as in \eqn{eq:PSD-noise}.
When using such a colored-noise to sample action-sequences inside CEM  (more details below), we obtain a dramatically improved speed and performance.
Considering the spectrum of the successful action sequences found by our proposed \Method{} method (green lines in \fig{fig:noise-psd}(b)) we see a clear preference of low frequencies as well as a sharp drop for the highest frequency (corresponding to alternating actions at every step). Regardless of whether we use $\beta=2$ or $\beta=4$, the action sequence follows roughly $\beta=1.5$ with an additional bump at $2$-$3$ Hz.

We introduce the colored-noise in CEM as a function of $\beta$ which creates correlated action sequences with a PSD as in \eqref{eq:PSD-noise}. For sampling, we use the efficient implementation of \cite{Timmer1995generatingCN} based on Fast Fourier Transform \cite{cochran1967fft}.
It relies on the fact that PSD of a time-series can be directly modified in the frequency space. Indeed, if we want to sample actions with a PSD as in \eqref{eq:PSD-noise},  we have to apply the following transformation to the original white noise actions $a(t)$:
\begin{align*}
\overline{a}(t) = \mathcal{F}^{-1}\left[\frac{1}{f^{\beta/2}} \mathcal{F}[a(t)]\right] \quad \textrm{gives} \quad \textrm{PSD}_{\overline{a}}(f) = \left\|\frac{1}{f^{\beta/2}} \mathcal{F}[a(t)](f)  \right\|^2 = \frac{1}{f^{\beta}}\textrm{PSD}_a(f) \propto \frac{1}{f^\beta}
\end{align*}
The resulting sampling function $\mathcal{C}^\beta(d, h)$ returns $d$ (one for each action dimension) sequences of length $h$ (horizon) sampled from colored noise distribution with exponent $\beta$ and with zero mean and unit variance.



\subsection{CEM with memory}
In the standard CEM, once the inner loop is completed, the optimized Gaussian distribution and the entirety of all the elite-sets generated at each iteration get discarded. According to the parameters used in \citet{chua2018pets}, this amounts to an average of $\sim 55000$ discarded actions \textbf{per step}.
To increase efficiency, the following improvements reuse some of this information:

\textbf{1. \emph{Keep} elites:}  Storing the elite-set generated at each inner CEM-iteration and adding a small fraction of them to the pool of the next iteration, instead of discarding the elite-set in each CEM-iteration.

\textbf{2. \emph{Shift} elites:} Storing a small fraction of the elite-set of the last CEM-iteration and add each a random action at the end to use it in the next environment step.

The reason for not shifting the entire elite-set in both cases is that it would shrink the variance of CEM drastically in the first CEM-iteration because the last elites are quite likely dominating the new samples and have small variance. We use a fraction of 0.3 in all experiments.

\subsection{Smaller Improvements}
\textbf{Executing the best action (\emph{best-a})} \quad
The purpose of the original CEM algorithm is to estimate an unknown probability distribution.
Using CEM as a trajectory optimizer detaches it from its original purpose.
In the MPC context we are interested in the best possible action to be executed.
For this reason, we choose the first action of the best seen action sequence,
rather than executing the first mean action, which was actually never evaluated.
Consequently, we add the mean to the samples of the last CEM-iteration
to allow the algorithm to still execute the mean action. For more details, see \sec{sec:sup:add-mean}.


\textbf{Clipping at the action boundaries (\emph{clip})}\quad
Instead of sampling from a truncated normal distribution, we sample from the unmodified normal distribution (or colored-noise distribution) and clip the results to lie inside the permitted action interval. This allows to sample maximal actions more frequently.

\textbf{Decay of population size (\emph{decay})}\quad
One of the advantages of CEM over the simplest Evolution Strategies is that the standard deviation is not fixed during the optimization procedure, but adapts according to the elite-set statistics.
When we are close to an optimum, the standard deviation will automatically decrease, narrowing down the search and fine-tuning the solution.
For this reason, it is sufficient to sample fewer action sequences as the CEM-iterations proceed. We introduce then an exponential decrease in population size of a fixed factor $\gamma$.
The population size of iteration $i$ is $N_i=\max(N \gamma^{-i}, 2K)$,
where the $\max$ ensures that the population size is at least double the size of the \textit{elite-set}.


The final version of the algorithm is showed in Alg.~\ref{algo:improved-cem}.
Hyper-parameters are given in \sec{sec:sup:params} in the supplementary. Except $\beta$ we use the same parameters for all settings. The planning horizon is 30.

\newcommand{\newMPC}[1]{{\color{ourblue!80!black} #1}}
\newcommand{\newiCEM}[1]{{\color{ourorange!70!black} #1}}
\newcommand\mycommfont[1]{\footnotesize\textcolor{gray}{#1}}
\SetCommentSty{mycommfont}

\begin{algorithm}
  Parameters:

  \quad $N$: number of samples; $h$: planning horizon; $K$: size of elite-set; $\beta$: colored-noise exponent\\
  \quad \textit{CEM-iterations}: number of iterations; $\gamma$: reduction factor of samples, $\sigma_{init}$: noise strength

  \For{t = 0 \textbf{to} T$-1$}{
    \eIf{t == 0}{
      $\mu_0$ $\leftarrow$ constant vector in $\mathbb{R}^{d\times h}$
    }{
      \newMPC{$\mu_t$ $\leftarrow$ shifted $\mu_{t-1}$ (and repeat last time-step)}
    }
    $\sigma_t$ $\leftarrow$ constant vector in $\mathbb{R}^{d\times h}$ with values $\sigma_{init}$

    \For{i = 0 \textbf{to} CEM-iterations$-1$}{
      \newiCEM{$N_i$ $\leftarrow$ $\max(N \cdot \gamma^{-i}, 2\cdot K)$}

      samples $\leftarrow N \textrm{ samples from } \mathcal{N}(\mu_t,\textrm{diag}(\sigma_t^2)) $  \tcp*{only CEM \& \CEMMPC}

      \newiCEM{samples $\leftarrow$ $N_i$ samples from clip$(\mu_t + \mathcal{C}^\beta(d, h) \odot \sigma_t^2)$} \tcp*{only \Method}

      \newiCEM{\eIf{i == 0}{
        add fraction of \textbf{shifted} elite-set$_{t-1}$ to samples
      }{
        add fraction of elite-set$_t$ to samples
      }

       \If{ i == last-iter}{
         add mean to samples
       }
       }

      costs $\leftarrow$ cost function $f(x)$ for $x$ in samples

      elite-set$_t$ $\leftarrow$ best $K$ samples according to costs

      $\mu_t$, $\sigma_t$ $\leftarrow$ fit Gaussian distribution to elite-set$_t$ \newMPC{with momentum}
    }

    execute action in first $\mu_t$ \tcp*{only CEM and \CEMMPC}

    \newiCEM{execute first action of best elite sequence} \tcp*{only \Method}

  }
  \caption{Proposed \Method{} algorithm. Color \newiCEM{brown is \Method{}} and
    \newMPC{blue is \CEMMPC{} and \Method}.}
 \label{algo:improved-cem}
\end{algorithm}

\section{Experiments}\label{sec:results}
The aim of the experiment section is to benchmark CEM-based methods on hard high-dimensional robotic tasks that need long horizon planning and study their behavior in the low-sampling budget regime.
The control tasks range from locomotion to manipulation with observation-dimension ranging from 18 to 376, and action-spaces up to 30 dimensions.
We use the ground truth dynamics model given by the Mujoco simulator
 as well as learned latent-dynamics models in the PlaNet~\cite{hafner2018planet} framework. Details and videos can be found in the Supplementary.

\subsection{Environments}\label{sec:environments}
First, we consider the following three challenging the environments contained in OpenAI Gym \cite{brockman2016openai}:

\textbf{\HCR{}:} (Gym v3) A half-cheetah agent should maximize its velocity in the positive x-direction. In contrast to the standard setting, we prohibit a rolling motion of the cheetah, commonly found by strong optimization schemes, by heavily penalizing large angles of the root joint.

\textbf{\HS{}:} (Gym v2) A humanoid robot is initialized in a laying position, see \fig{fig:environments}. The goal is to stand-up without falling, \ie reaching as high as possible with the head.

\textbf{\FPP{} (sparse reward):} (Gym v1) A robotic manipulator has to move a box, randomly placed on a table, to a randomly selected target location. The agent is a Cartesian coordinate robotic arm with a two finger gripper attached to its end effector. The reward is only the negative Euclidean distance between box and target location, so without moving the box there is no reward.

Furthermore, we test \Method{} on three environments from the DAPG\footnote{\label{dapg_repo}\url{https://github.com/aravindr93/hand_dapg}} project~\cite{Rajeswaran-RSS-18}.
The basis of these environments is a simulated 24 degrees of freedom ShadowHand. Each environment requires the agent to solve a single task:

\textbf{\Door{}:} The task is to open a door by first pushing down the door handle which releases the latch, enabling the agent to open the door by pulling at the handle.
The reward (as in ~\cite{Rajeswaran-RSS-18}) is the sum of the negative distance between palm and door handle,
the openness of the door and a quadratic penalty on the velocities.
Additional bounties are given for opening the door.
The state space contains the relative joint positions of the hand, the latch position, the absolute door, palm and handle position, the relative position between palm and handle and a flag indicating whether the door is open or not. 

\textbf{\DoorNS{}:} The same as \Door{} except the reward does not contain the distance of the palm to the handle, so without opening the door there is no reward.

\textbf{\Reloc{}:} In the relocate environment the task, see \fig{fig:environments}, is to move a ball to a target location. To achieve the goal, the ball needs to be lifted into the air. The reward signal is the negative distance between palm and ball, ball and target and bounties for lifting up the object and for when the object is close to the target. The state space contains the relative joint positions of the hand and the pairwise relative positions of the palm, the ball, and the target. 

\subsection{Main results}
\begin{figure}
  \centering
  \begin{tabular}{ccc}
    \HCR{} & \HS{} & \FPP{}\\
    \includegraphics[height=0.15\linewidth]{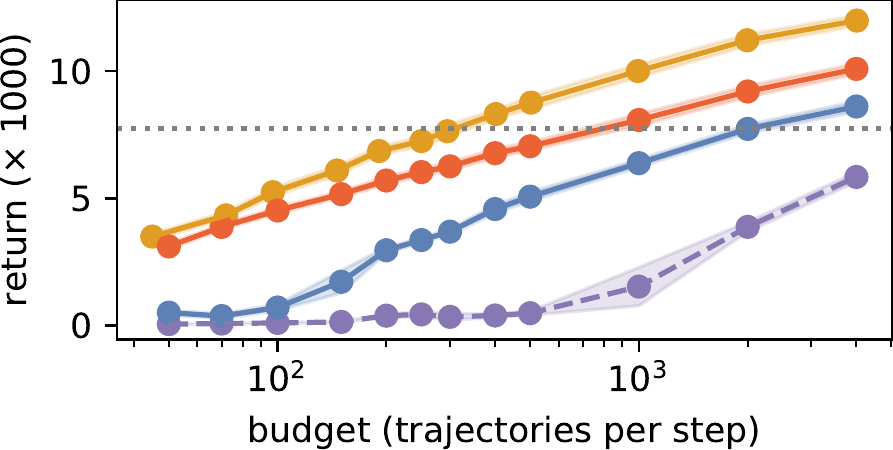}&
    \includegraphics[height=0.15\linewidth]{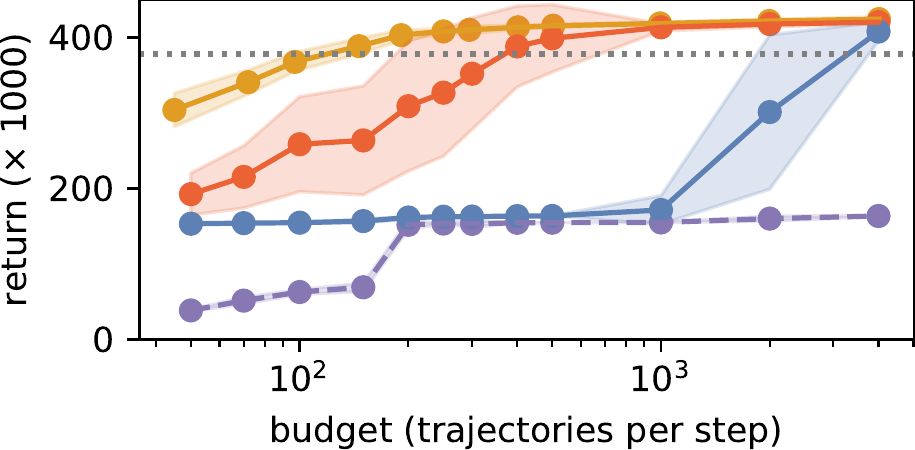}&
    \includegraphics[height=0.15\linewidth]{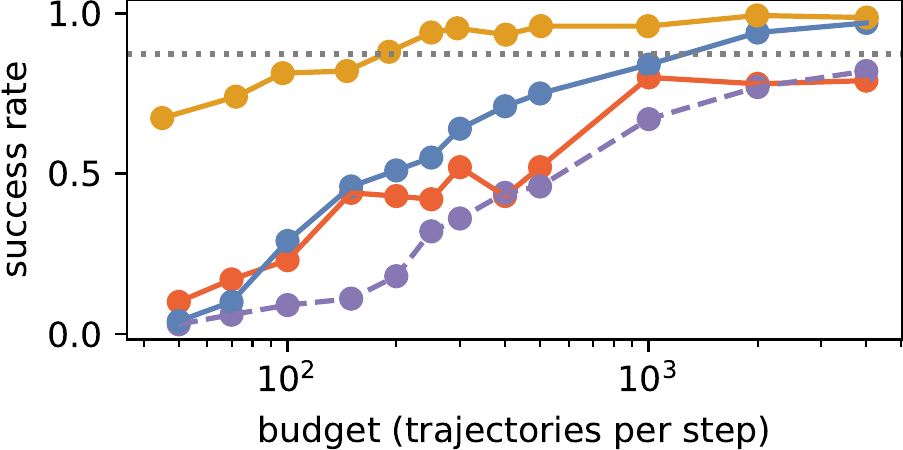}\\
    \Door & \DoorNS & \Reloc\\
    \includegraphics[height=0.15\linewidth]{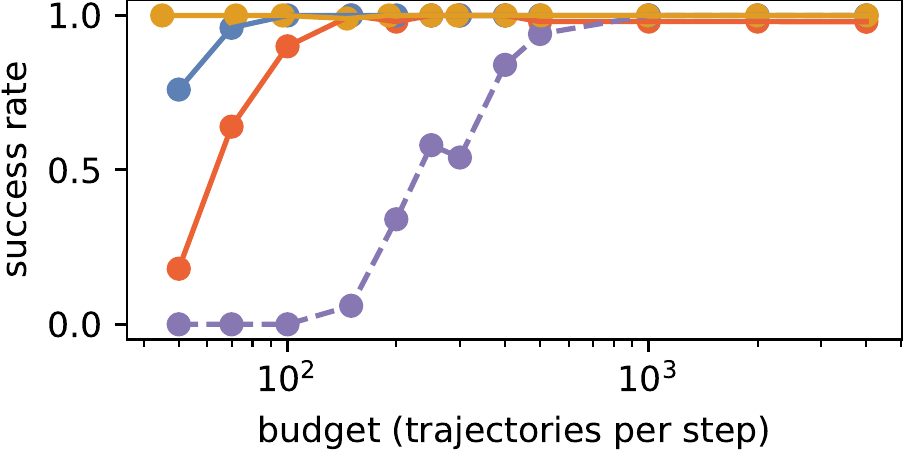}&
    \includegraphics[height=0.15\linewidth]{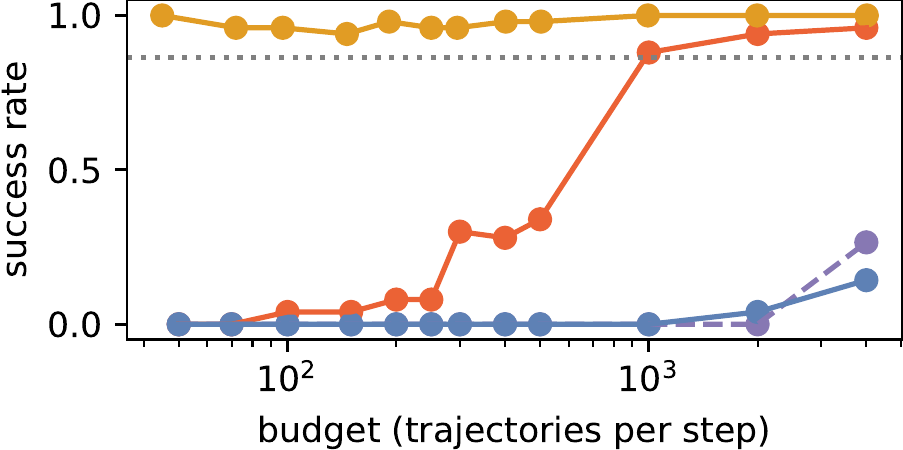}&
    \includegraphics[height=0.15\linewidth]{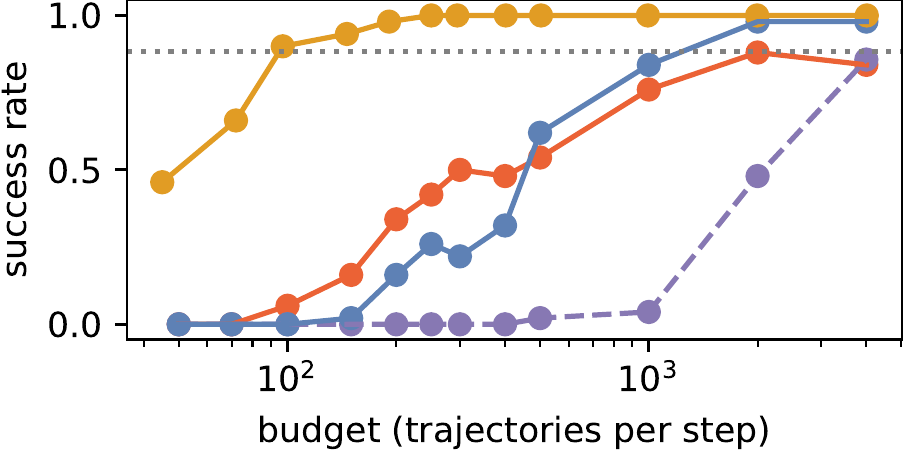}
  \end{tabular}\\
  {\small \textcolor{ourorange}{\rule[2pt]{20pt}{1.5pt}} \Method{} \quad \textcolor{ourred}{\rule[2pt]{20pt}{1.5pt}} \CEMPETS{} \quad \textcolor{ourblue}{\rule[2pt]{20pt}{1.5pt}} \CEMMPC{} \quad \textcolor{ourviolet}{\rule[2pt]{4pt}{1.5pt}\ \rule[2pt]{4pt}{1.5pt}\ \rule[2pt]{4pt}{1.5pt}} CEM}
  \caption{Performance dependence on the planning budget. Notice the log-scale on the $x$-axis.}
  \label{fig:budget}
\end{figure}

We want to obtain a sample efficient CEM that can potentially be used in real-time given a moderate model runtime.
For this reason, we study how the performance degrades when decreasing the number of samples per time-step,
in order to find a good compromise between execution speed and desired outcome.

\Fig{fig:budget} presents the performance of \Method{}, \CEMMPC{}, \CEMPETS{} and vanilla CEM for different budgets, where a budget is the total number of trajectories per \emph{step}.
It clearly demonstrates that \Method{} is the only method to perform well even with extremely low budgets. In addition, \Method{} has consistently higher performance than the baselines for all considered budgets, see also \tab{tab:performance}.

\begin{table}[b]
  \caption{Sample efficiency and performance increase of \Method{} \wrt the best baseline.
    The first 4 columns consider the budget needed to reach 90\% of the best baseline (dashed lines in \fig{fig:budget}). The last column is the average improvement over the best baseline in the given budget interval.}
  \small
  \begin{center}
    \makebox[0pt]{\begin{tabular}{@{}rrrrr|cr@{}}
      \toprule
                & 90\% base- & $\sim$ budget & $\sim$ budget & efficiency & \multicolumn{2}{c}{\Method{} \wrt baseline}\\
              & line@4000 & \Method{} & baseline & factor & budgets   & \%\quad\      \\
      \midrule
      \HCR    & 7744      & 312       & 840      & 2.7    & 50--1000  & 120\%  \\  
      \HS     & 378577    & 121       & 372      & 3.06   & 50--1000  & 128\%  \\ 
      \FPP    & 0.87      & 185       & 1330     & 7.2    & 50--1000  & 243\%  \\ 
      \DoorNS & 0.86      & 45        & 985      & 21.9   & 100--1000 & 1030\% \\ 
      \Reloc  & 0.88      & 95        & 1300     & 13.7   & 100--1000 & 413\%  \\ 
        \bottomrule
    \end{tabular}}
  \end{center}
  \label{tab:perform:compare}
\end{table}
To quantify the improvements, \tab{tab:perform:compare} compares \Method{} with the respective best baseline
in each environment.
We report the sample efficiency factor based on the approximate budget needed to reach 90\% of the best baseline performance (at budget 4000) and see that \Method{} is 2.7-21.9$\times$ more sample efficient.
Similarly, we consider how much higher performance \Method{} has \wrt the best baseline for a given budget (averaged over budgets$<1000$) and find 120-1030\% of the best baseline performance.

\begin{figure}
  \centering
  \begin{tabular}{c@{\ \ }c@{\ \ }c}
    \env{Cheetah Run} & \env{Walker Walk} & \env{Cup Catch}\\
    \includegraphics[width=.31\linewidth]{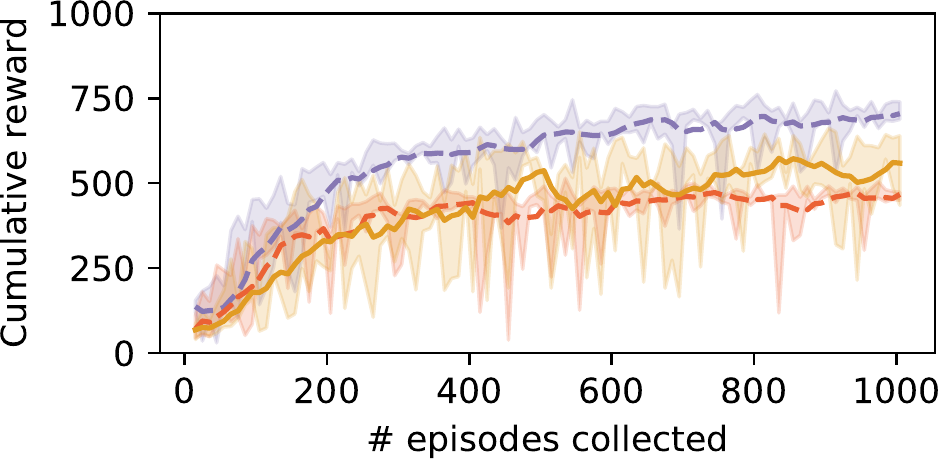}&
    \includegraphics[width=.31\linewidth]{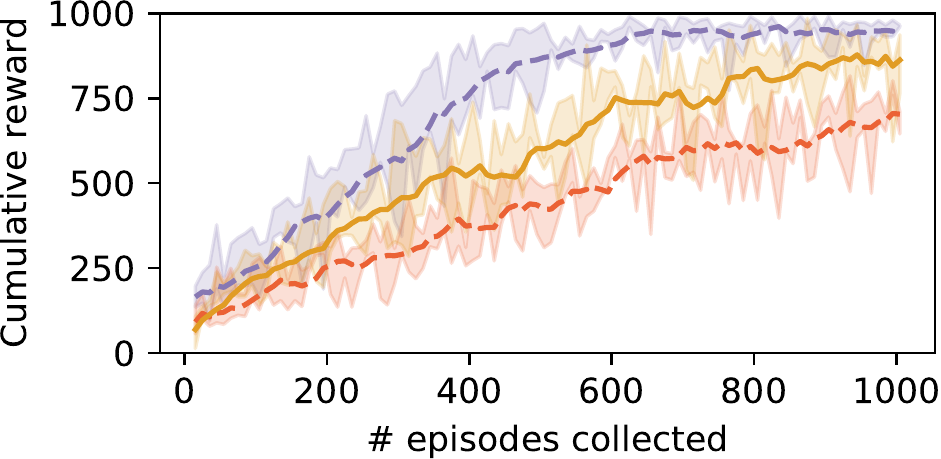}&
    \includegraphics[width=.31\linewidth]{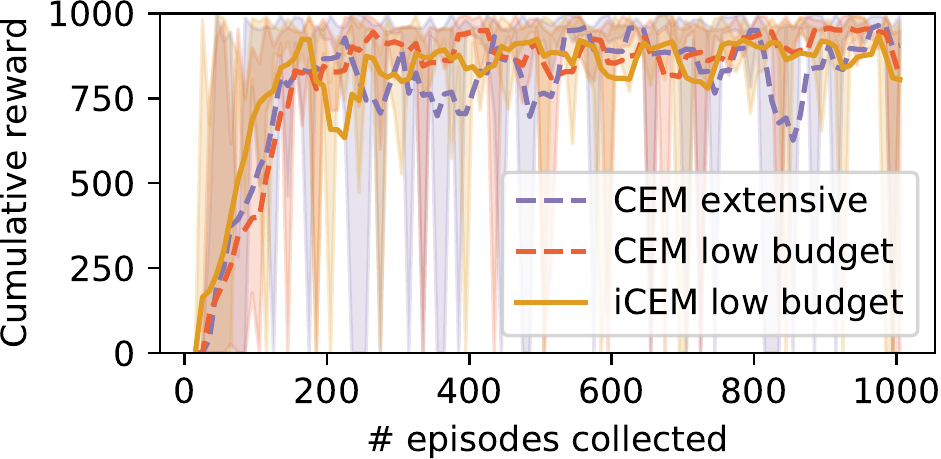}
  \end{tabular}
  \caption{PlaNet performance with low budget.
    Task performance for planning using an extensive CEM variant (budget 10000, 10 CEM-iterations) and two low-budget variants of CEM and \Method{} (budget 366, 3 CEM-iterations).
    The dynamics models used for planning are learned from pixels with PlaNet \cite{hafner2018planet}
    with training data repeatedly collected with the respective planner.
    Shown is the mean and min/max-band cumulative reward (three independent restarts) with average-smoothing over 5 test collections (50 episodes).
   \Method{} outperforms the low-budget baseline on \env{Cheetah Run} and \env{Walker Walk}, and demonstrates similar performance on \env{Cup Catch}.
}
  \label{fig:planet}
\end{figure}

\textbf{Planning using learned dynamics models:}
In addition to planning in environments with given ground-truth dynamics,
we investigate the behavior of \Method{} for planning using learned dynamics models.
For this, we train dynamics models from pixel input on several DeepMind control suite tasks using PlaNet \cite{hafner2018planet}.
We compare the performance of the entire training and planning process, see Fig.~\ref{fig:planet}, with
 (a) the CEM planner with budget 10000 and 10 CEM-iterations, (b) \Method{} with small budget (366) and 3 CEM-iterations, and (c) the CEM planner with small budget (366) and 3 CEM-iterations.
For all planners, we execute the best trajectory action.
We observe that \Method{} with a budget of only 366 is not far behind the extensive CEM (a).
Moreover, \Method{} is clearly better than the baseline (c) with the same low budget for the \env{Cheetah Run} and \env{Walker Walk} environments.
\env{Cup Catch} is a challenging learning task due to its sparse reward.
Presumably, training progress largely depends on observing successful rollouts early in training.
On this task, \Method{} reaches similar performance to the other CEM methods.
We provide further details and results in the supplementary material.

\textbf{Towards real-time control:}
Using the learned models in the PlaNet approach we reach real-time planning with \Method{} using our
own PyTorch implementation, see \tab{tab:runtimes}.
Also with the ground-truth models and CPU-parallelization we reach close to real-time performance for
simple environments (\env{HalfCheetah}).
\begin{table}
  \caption{Runtimes for \Method{} with different compute budgets using Mujoco simulator and the PlaNet models.
    Times are given in seconds per env-step on $^1$: Xeon(R) Gold 6154 CPU @ 3.00GHz, and \newline $^2$: Xeon$^\textrm{\textregistered}$ Gold 5220, NVidia$^\textrm{\textregistered}$ Quadro RTX 6000.}
    \small
  \begin{center}
    \begin{tabular}{@{}lrrrrrl@{}}
    \toprule
      &  & \multicolumn{4}{c}{Budget (trajectories per step)} & \\\cline{3-6}\Tstrut
      Envs                  & Threads & 100   & 300    & 500    & 2000    & dt                     \\
      \midrule
      \multirow{2}{*}{\HCR$^1$} & 1     & 0.326 & 0.884  & 1.520  & 5.851   & \multirow{2}{*}{0.05}  \\
                            & 32    & 0.027 & 0.066 & 0.109	&0.399  &                        \\[.15em]
      \multirow{2}{*}{\HS$^1$}  & 1     & 2.745 & 8.811  & 13.259 & 47.469  & \multirow{2}{*}{0.015} \\
                            & 32        & 0.163 & 0.456	& 0.719  &	2.79 \\[.15em]
      \multirow{2}{*}{\FPP$^1$} & 1     & 8.391 & 26.988 & 40.630 & 166.223 & \multirow{2}{*}{0.04}  \\
                            & 32       & 0.368	&1.068 &	1.573   & 6.010 &                        \\
      \midrule
                            & \multicolumn{3}{c}{\Method{} (366)} & \multicolumn{2}{c}{CEM (10000)}  & dt                     \\
      \cline{2-7}\Tstrut
      PlaNet (PyTorch)$^2$ & \multicolumn{3}{c}{0.044$\rpm$0.003} & \multicolumn{2}{c}{0.18$\rpm$0.031} & 0.04--0.08\\
      \bottomrule
    \end{tabular}
  \end{center}
  \label{tab:runtimes}
\end{table}

\subsection{Ablation study}\label{sec:ablations}

\begin{figure}[tb]
  \centering
  \begin{tabular}{c@{\ \ \ }c@{\ \ \ }c}
    HalfCheetah (Running) 100 & Fetch Pick and place (300)&Relocate (300)\\
    \includegraphics[width=.31\linewidth]{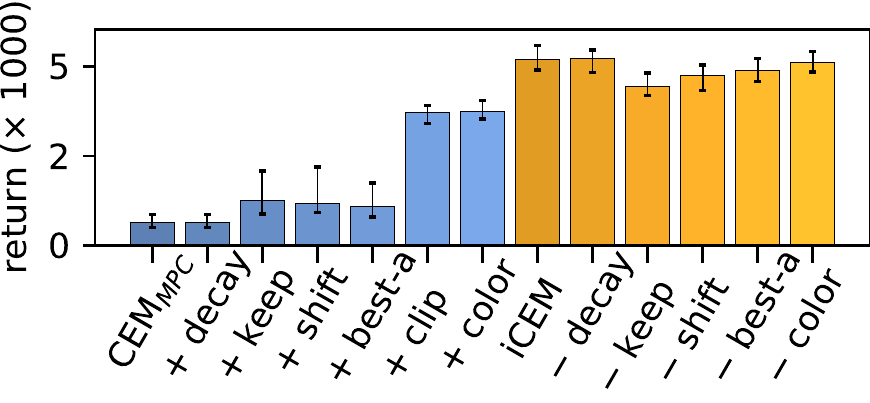}&
    \includegraphics[width=.31\linewidth]{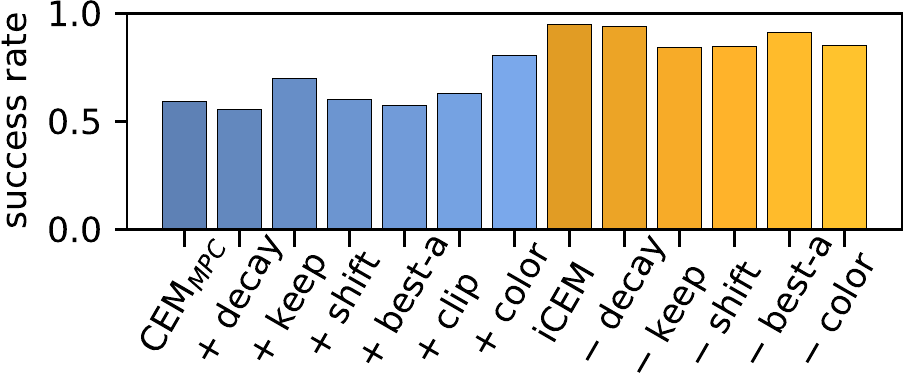}&
    \includegraphics[width=.31\linewidth]{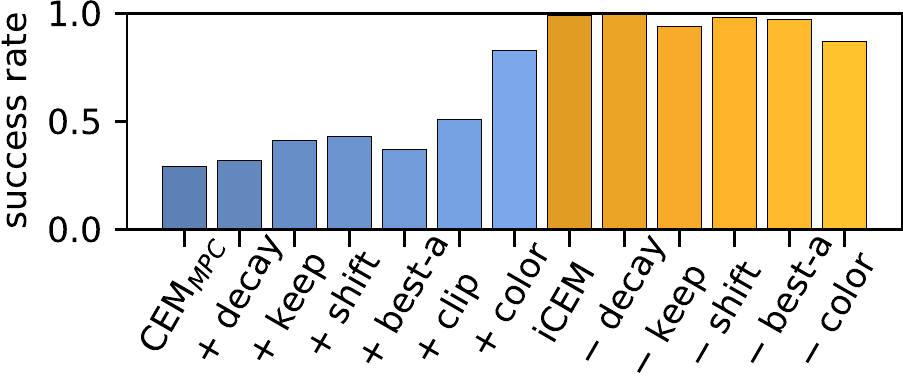}
  \end{tabular}
  \caption{Ablation studies. Blue bars show \CEMMPC{} with each improvement added separately.
    Yellow bars show \Method{} with each features removed separately. Feature names are listed in \sec{sec:methods}.
  }
  \label{fig:ablations:select}
\end{figure}

To study the impact of each of our improvement individually, we conducted ablations of \Method{} (orange bars in \fig{fig:ablations:select}) and additions to \CEMMPC{} (blue bars in \fig{fig:ablations:select}) for some environments and budgets, see \sec{sec:sup:ablations} for all combinations and more details.

Some components have bigger individual impact than others, \eg using \emph{color}ed  noise consistently has a huge impact on the final result followed by \emph{keep} and \emph{shift} elites and \emph{best-a}ction execution.
However, the addition of all components together is necessary to reach top performance.
As expected, the impact of the different additions become more relevant in the low-budget regime.

\section{Conclusions}\label{sec:conclusions}
In this work, we introduced \Method{}: a sample-efficient improvement of CEM intended for real-time planning.
Most notably, we introduce temporally correlated action sampling and memory for previous trajectories. These additions were crucial for solving for the first time complicated tasks in MBRL with \emph{very few} samples, e.g., humanoid stand-up or door opening (with sparse rewards) with only 45 trajectories per step.

With this budget, we manage to enter in the real-time regime, as we saw from the experiments with learned models.
We hope this encourages future work with zero-order optimizers for real-time robot control.

\clearpage
\section*{\centering \LARGE Supplementary Material}
\beginsupplement
\thispagestyle{empty}
\pagestyle{empty}

\appendix

In this supplementary material we detail the performances of \Method{} with both ground truth and learned models, and discuss the hyperparameter selection with a sensitivity analysis. We present the ablation figures for all the environments and 3 fixed budgets. We conclude with an analysis stressing the relation between time-correlated action sequences and their power spectrum. Some videos of \Method{} in action can be found at \url{https://sample-efficient-cross-entropy.github.io/}.



\section{Pseudocode of the vanilla Cross Entropy Method (CEM) in the MPC setting.}\label{sec:sup:CEM}

\begin{algorithm}[H]
  \SetAlgoRefName{S1}
  Parameters:

  \quad $N$: number of samples; $K$: size of elite-set; $h$: horizon;\\
  \quad $\sigma_{init}$: initial standard deviation;
  \emph{CEM-iterations}: number of iterations

  \For{t = 0 \textbf{to} T$-1$}{
    $\mu_0$ $\leftarrow$ zeros in $\mathbb{R}^{d \times h}$

    $\sigma_0$ $\leftarrow$ constant vector in $\mathbb{R}^{d \times h}$ with values $\sigma_{init}$

    \For{i = 0 \textbf{to} CEM-iterations$-1$}{ \label{algo:cem:inner}



      samples $\leftarrow N$ samples from $  \mathcal{N}(\mu_t,\textrm{diag}(\sigma_t^2))$

      costs $\leftarrow$ cost function $f(x)$ for $x$ in samples

      elite-set $\leftarrow$ best $K$ samples according to costs

      $\mu_t$, $\sigma_t$ $\leftarrow$ fit Gaussian distribution to elite-set \label{algo:cem:refit}
    }\label{algo:cem:inner:end}
    execute first action of mean sequence $\mu_t$
  }
  \caption{Cross-Entropy Method (CEM) for Trajectory Optimization}
  \label{algo:cem}
\end{algorithm}

\section{Performance results}\label{sec:sup:results}
\begin{figure}[h]
  \centering
  \begin{tabular}{c@{\ \ }c@{\ \ }c}
    \env{Reacher Easy} & \env{Finger Spin} & \env{Cartpole Swingup}\\
    \includegraphics[width=.31\linewidth]{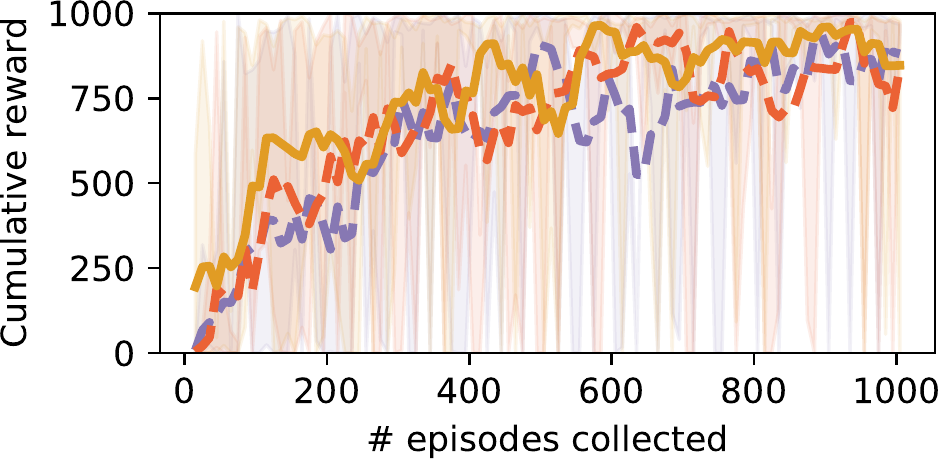}&
    \includegraphics[width=.31\linewidth]{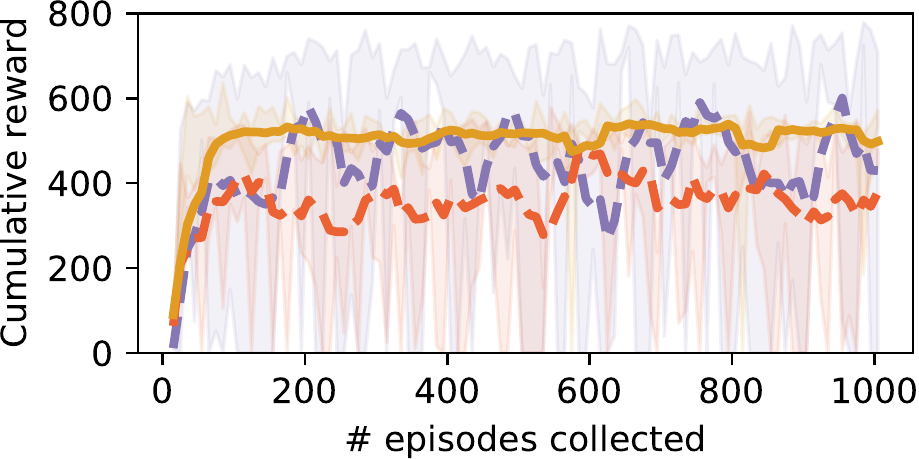}&
    \includegraphics[width=.31\linewidth]{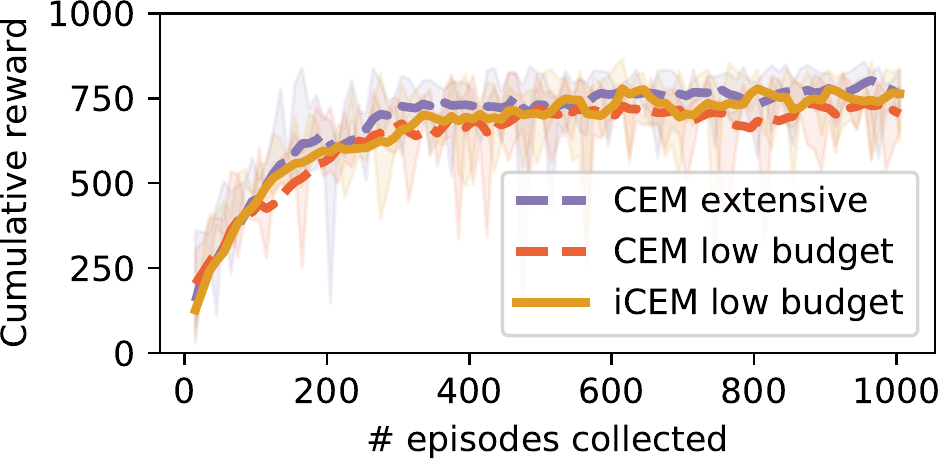} \\
  \end{tabular}
  \caption{Additional PlaNet experiments. For details, see Fig.~\ref{fig:planet}.}
  \label{fig:planet_additional}
\end{figure}

\tab{tab:performance} shows the performance values for a selection of budgets in all environments.
The values are reported for $50$ independent runs (and $100$ for \FPP{}) in the case of the ground-truth environments. 
For the PlaNet experiments we report the statistics for $3$ independent training runs with $10$ evaluation rollouts each. 
Note that for the success rate the variance is defined by the rate itself (Bernoulli distribution).     
\tab{tab:performance} is complemented by \fig{fig:planet_additional}, which shows the additional PlaNet experiments with \env{Reacher Easy}, \env{Finger Spin} and \env{Cartpole Swingup}.

\begin{table*}[h]
  \centering
  \caption{Performances for all environments for a selection of budgets. We report the cumulative reward (marked with $^1$) and the success rate (marked with $^2$).}
  \begin{adjustbox}{max width=\textwidth}
	\begin{tabular}{@{}l@{\hskip1em} *{9}{L} @{\hskip1em} c <{\clearrow}@{}}
         \toprule
         \multicolumn{1}{c}{Envs}  & \multicolumn{2}{c}{Budget 100}  && \multicolumn{2}{c}{Budget 300}  && \multicolumn{2}{c}{Budget 500}                                                         \\
         \cline{2-3} \cline{5-6} \cline{8-9}\Tstrut
         \hfill                            & \Method{}               & \CEMMPC                 && \Method          & \CEMMPC            && \Method{}          & \CEMMPC            \\
         \midrule
         \HCR$^1$         & \textbf{5236$\rpm$167}           & 699$\rpm$120            && \textbf{7633$\rpm$250}    & 3682$\rpm$119      && \textbf{8756$\rpm$255}      & 5059$\rpm$179        \\
         \HS$^1$            & \textbf{368\,k$\rpm$12\,k}       & 155\,k$\rpm$488         && \textbf{411\,k$\rpm$5\,k} & 163\,k$\rpm$495    && \textbf{416\,k$\rpm$1.8\,k} & 164\,k$\rpm$158    \\
         \FPP$^2$           & \textbf{0.81}          & 0.29          && \textbf{0.95}   & 0.64     && \textbf{0.96}      & 0.75      \\
         \Door$^2$                        & \textbf{1.0}            & \textbf{1.0}              && \textbf{1.0}      & \textbf{1.0}        && \textbf{1.0}         & \textbf{1.0}       \\
         \DoorNS$^2$        & \textbf{0.96}           & 0.0            && \textbf{0.96}    & 0.0       && \textbf{0.98}     & 0.0      \\
         \Reloc$^2$                    & \textbf{0.9}            & 0.0            && \textbf{1.0}     & 0.22     && \textbf{1.0}       & 0.62     \\
      \midrule
       & \multicolumn{2}{l}{iCEM (366)} & & \multicolumn{2}{l}{plain CEM (366)} & & \multicolumn{2}{l}{plain CEM (10000)}\\
         \cline{2-9}\Tstrut
         PlaNet \env{Cheetah run}$^1$  & \multicolumn{2}{l}{531.65$\rpm$134.61} & &  \multicolumn{2}{l}{447.85$\rpm$44.98} & & \multicolumn{2}{l}{\textbf{679.85$\rpm$102.41}} \\
         PlaNet \env{Cup catch}$^1$  & \multicolumn{2}{l}{927.10$\rpm$47.71} & &  \multicolumn{2}{l}{760.57$\rpm$387.87} & & \multicolumn{2}{l}{\textbf{964.57$\rpm$18.41}} \\
         PlaNet \env{Walker walk}$^1$  & \multicolumn{2}{l}{865.30$\rpm$67.11} & &  \multicolumn{2}{l}{710.09$\rpm$78.29} & & \multicolumn{2}{l}{\textbf{940.62$\rpm$24.79}}\\
         PlaNet \env{Reacher Easy}$^1$  & \multicolumn{2}{l}{\textbf{905.17$\rpm$247.42}} & &  \multicolumn{2}{l}{879.20$\rpm$246.46} & & \multicolumn{2}{l}{831.70$\rpm$338.28}\\
         PlaNet \env{Finger Spin}$^1$  & \multicolumn{2}{l}{\textbf{539.30$\rpm$27.95}} & &  \multicolumn{2}{l}{437.33$\rpm$132.69} & & \multicolumn{2}{l}{360.67$\rpm$248.75}\\
         PlaNet \env{Cartpole Swingup}$^1$  & \multicolumn{2}{l}{\textbf{772.52$\rpm$44.08}} & &  \multicolumn{2}{l}{702.2$\rpm$91.87} & & \multicolumn{2}{l}{768.01$\rpm$44.61}\\
         \bottomrule
       \end{tabular}
  \end{adjustbox}
  \label{tab:performance}
\end{table*}

\subsection{Budget selection}
\tab{tab:budget} gives different budgets used for evaluating \Method{} performance and the corresponding internal optimizer settings. Note that the number of \emph{CEM-iterations} and the number of initial trajectories $N$ depends on the overall budget and, due to the decay $\gamma=1.25$, there are more \emph{CEM-iterations} possible for \Method{} while keeping the same budget.

\begin{table}[h]
\centering
\caption{Budget-dependent internal optimizer settings (notation: \emph{CEM-iterations} / $N$). }
\begin{adjustbox}{max width=\textwidth}
\begin{tabular}{@{}l *{12}{C}@{}}
\toprule
 & \multicolumn{12}{C}{Budgets} \\
 \cline{2-13} \Tstrut
  & 50 & 70      & 100     & 150     & 200     & 250     & 300      & 400      & 500      & 1000     & 2000     & 4000     \\
  \midrule
  \Method &                2 / 25 & 2 / 40 & 3 / 40 & 3 / 60 & 4 / 65 & 4 / 85 & 4 / 100 & 5 / 120 & 5 / 150 & 6 / 270 & 8 / 480 & 10 / 900\\
  CEM     &  2 / 25        & 2 / 35 & 2 / 50 & 2 / 75 & 3 / 66 & 3 / 83 & 3 / 100 & 4 / 100 & 4 / 125 & 4 / 250 & 6 / 333 & 8 / 500 \\
\bottomrule
\end{tabular}
\end{adjustbox}
\label{tab:budget}
\end{table}

\section{Hyper-parameters}\label{sec:sup:params}
Zeroth order optimization requires minimal hyperparameter tuning in comparison to gradient descent methods,
to the extent that it is used itself to tune the hyperparameters of deep networks \cite{snoek2012nntuning}.

The main parameters in CEM, aside from the length of the planning horizon $h$ and the number of trajectories (determined by population size $N$ and number of \emph{CEM-iterations}), are:
the size of the \emph{elite-set} $K$,
the initial standard deviation $\sigma_{init}$ and the $\alpha$-momentum.

To these, \Method{} adds the colored-noise exponent $\beta$, the decay factor $\gamma$, and the fraction of reused elites $\xi$.
We unified the values of $\alpha, K, \sigma_{init}, \gamma$, and  $\xi$ for all the presented tasks, see \tab{tab:hyperparams:fixed}.

For experiments with the ground truth model we use an horizon of 30 and for
the PlaNet experiments we use the horizon of 12 to be consistent with the original PlaNet paper.
All the other parameters are fixed to the same values for all the environments, with the exception of the noise-exponent $\beta$, as reported in \tab{tab:hyperparams:env}.
\begin{table}[htbp]
\centering
\caption{Fixed Hyperparameters used for all experiments.}
\begin{tabular}{rllllll}
  \toprule
              & \# elites & initial std.& momentum & decay & fraction reused elites\\
              & $K$ & $\sigma_{init}$ & $\alpha$ &  $\gamma$ & $\xi$\\
  \midrule
  \Method                        & 10  & 0.5             & 0.1      & 1.25 & 0.3 \\
  CEM
                                   & 10  & 0.5             & --       & 1.0  & 0   \\
  \bottomrule
\end{tabular}
\label{tab:hyperparams:fixed}
\end{table}
\begin{table}[htbp]
\centering
\caption{Env-dependent Hyperparameter choices.}
\begin{tabular}{lll}
\toprule
    & \Method/CEM with ground truth & iCEM with PlaNet \\
  \midrule
  horizon $h$           & 30                        &  12                 \\
  \midrule
  colored-noise exponent $\beta$ & \begin{tabular}[t]{@{}ll@{}}
              0.25 & \HCR{}\\
              2.0 & \HS{}\\
              2.5 & \Door{} \\
              2.5 & \DoorNS \\
              3.0 & \FPP{} \\
              3.5 & \Reloc{}
            \end{tabular} & \begin{tabular}[t]{@{}ll@{}}
                              0.25 & \env{Cheetah run}\\
                              0.25 & \env{Cartpole Swingup} \\
                              2.5 & \env{Walker walk}\\
                              2.5 & \env{Cup Catch} \\
                              2.5 & \env{Reacher Easy} \\
                              2.5 & \env{Finger Spin} \\
                              \end{tabular}   \\
  \midrule
  initial std. $\sigma_{init}$ & & \begin{tabular}[t]{@{}ll@{}}
              0.5 & \env{Cheetah run}\\
              0.5 & \env{Walker walk}\\
              0.5 & \env{Cup Catch} \\
              0.5 & \env{Reacher Easy} \\
              1.0 & \env{Finger Spin} \\
              1.0 & \env{Cartpole Swingup} \\
              \end{tabular}   \\
\bottomrule
\end{tabular}
\label{tab:hyperparams:env}
\end{table}

\subsection{Choice of colored-noise exponent $\beta$}
The choice of the $\beta$ is intuitive and directly related to the nature of each task. Using colored-noise allows us to generate action sequences more specific to the intrinsic frequency of the considered tasks: higher $\beta$ for low-frequency control (\FPP{}, \Reloc{}, etc.) and lower $\beta$ for high-frequency control (\HCR{}), as shown in \tab{tab:hyperparams:env}.

For example, as we saw in \fig{fig:noise-psd}(b), the Humanoid prefers a non-flat spectral density with a predominance of lower frequencies, reason why feeding actions drawn from a Gaussian distribution would inevitably translate in a "waste" of energy.
By picking a $\beta$ in the right range, we avoid this and consequently make the whole optimization procedure more efficient.

Aside from this, providing a precise value for $\beta$ is not critical.
Environments that require a high-frequency control need a low $\beta$ otherwise a value around 2--4 seems adequate, as shown in the sensitivity plot in \fig{fig:sup:sensitivity:beta}.

\subsection{Sensitivity}\label{sec:sup:sensitivity}
\begin{figure}
  \centering
  \begin{tabular}{@{}cc@{}}
    (a) horizon $h$ &  (b) initial standard dev.\\
    \includegraphics[width=0.4\linewidth]{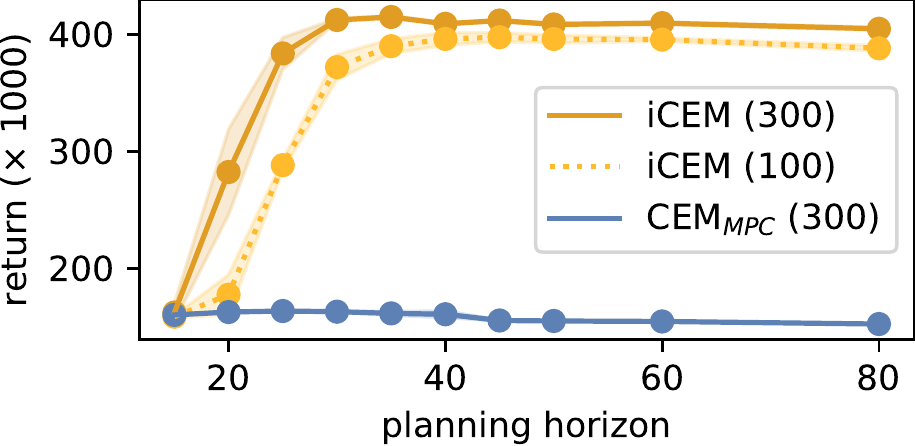}&
    \includegraphics[width=0.4\linewidth]{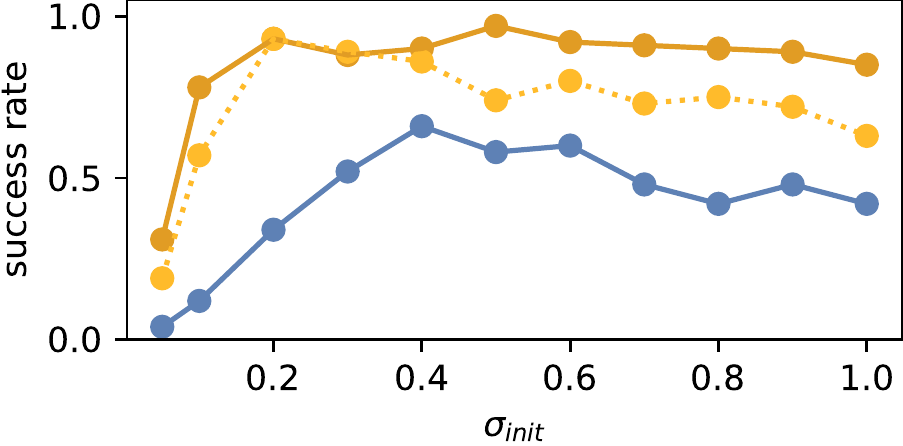}
  \end{tabular}
  \caption{Sensitivity to hyper-parameters of \Method{}. (a) horizon $h$ in \HS{} and in (b) the initial standard deviation for \FPP{}. See \fig{fig:sup:sensitivity:beta} for the sensitivity to $\beta$.  }
  \label{fig:sup:sensitivity}
\end{figure}

\begin{figure}
  \centering
  \begin{tabular}{@{}cc@{}}
    (a) \HCR & (b) \HS \\
    \includegraphics[width=0.4\linewidth]{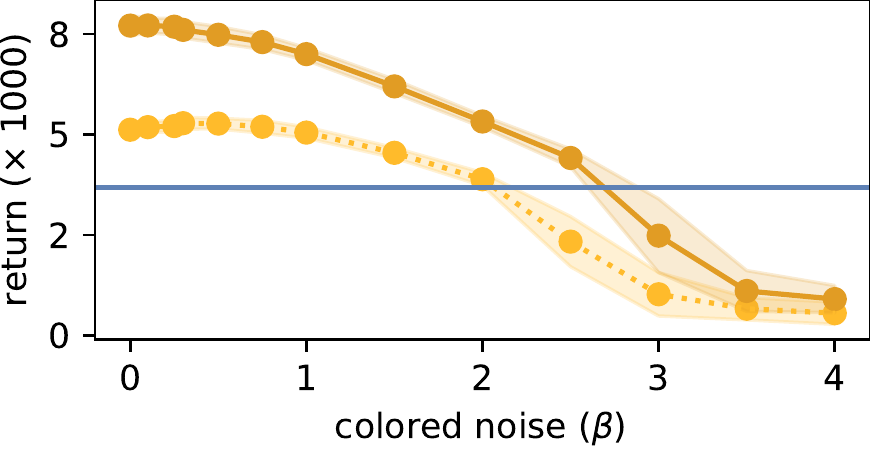}&
    \includegraphics[width=0.4\linewidth]{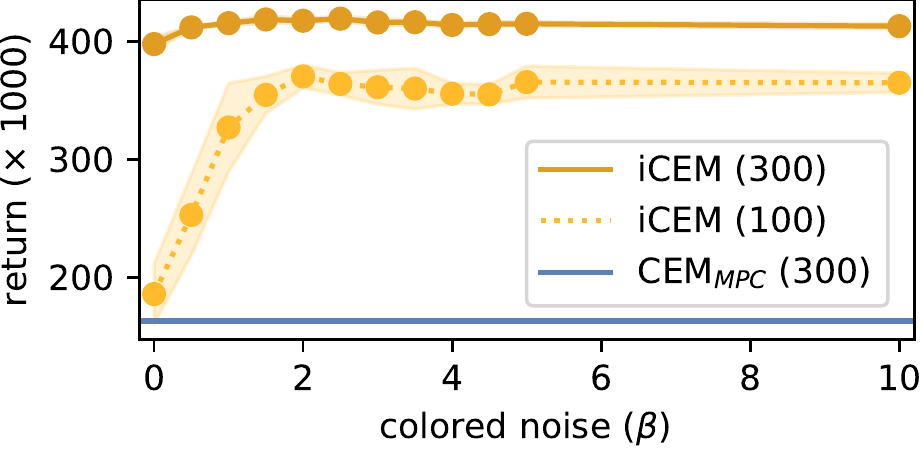}\\
    (c) \FPP & (d) \Reloc \\
    \includegraphics[width=0.4\linewidth]{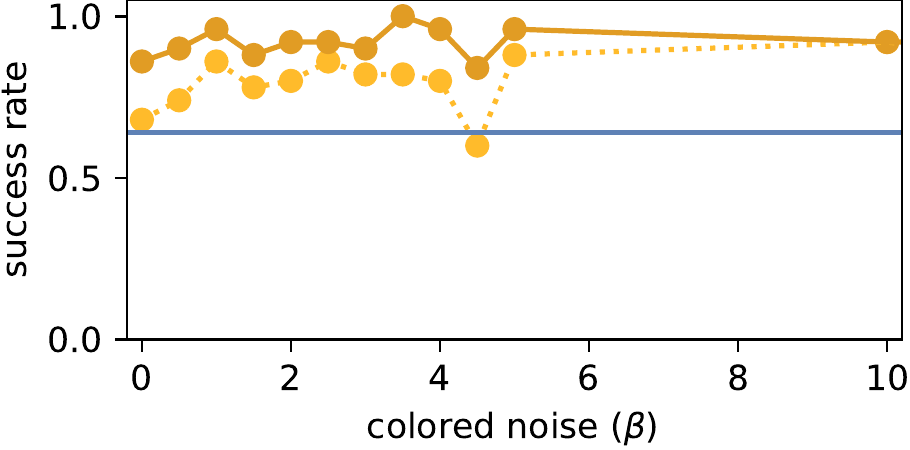}&
    \includegraphics[width=0.4\linewidth]{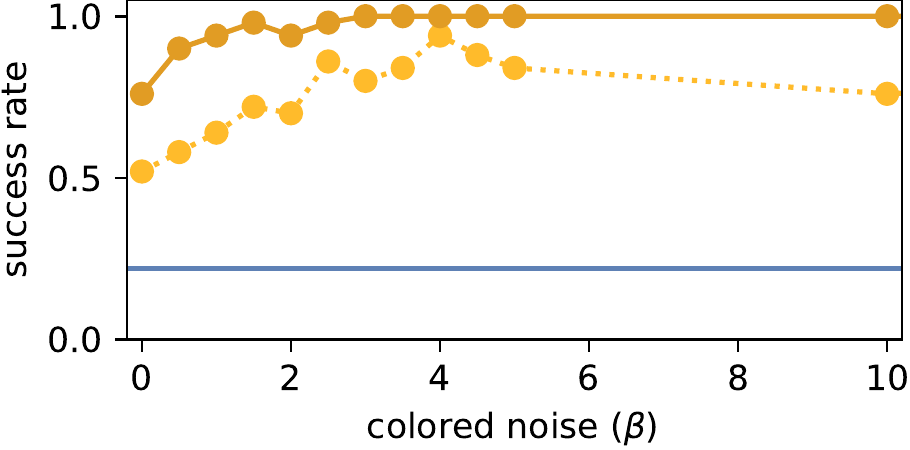}\\
  \end{tabular}
  \caption{Sensitivity to the colored noise exponent $\beta$ of \Method{}.}
  \label{fig:sup:sensitivity:beta}
\end{figure}

If the number of trajectories is high enough, there is little sensitivity to the other parameters, as shown in \fig{fig:sup:sensitivity} and \fig{fig:sup:sensitivity:beta}. This means that for very low budgets -- the ones relevant for real-time planning -- selecting the right parameters becomes more important.
We can the measure the impact of every parameter by comparing the first (low budget) and last column (higher budget) of \fig{fig:sup:ablations}. As the number of samples increases, adding features does not have significant consequences on the final performance.

However, selecting the colored-noise exponent $\beta$ in the right range, can still have a significant effect depending on the specific task, even for higher budgets.
For example, it is important to not use high values of $\beta$ for high-frequency control tasks like \HCR{}.
On the other hand, using higher $\beta$ on the \HS{} is fundamental when the provided budget is low (100). In fact, as illustrated in \fig{fig:sup:sensitivity:beta} (b), it is crucial to increase $\beta$ to any value above 2, in order to not sample uncorrelated action sequences. 

Besides that, \Method{} shows lower sensitivity with respect to the initial standard deviation  of the sampling distribution. As an example, we report the effect of $\sigma_{init}$ on the success rate of the \FPP{} task in \fig{fig:sup:sensitivity} (c): \CEMMPC{} prefers a narrower range for $\sigma_{init}$ between 0.4 and 0.6, in contrast to \Method{}, for which any value above 0.2 yields similar results.

Another relevant observation is the effect of the planning horizon length $h$ for the \HS{} in \fig{fig:sup:sensitivity} (a): even in the low-budget case, \Method{} can better exploit longer action sequences by generating samples with higher correlations in time.

\subsection{Hyperparameters for PlaNet}
We reimplemented PlaNet \cite{hafner2018planet} in PyTorch to conduct our experiments and borrow all algorithmic details and hyperparameter settings from the original paper. As in \cite{hafner2018planet}, for every training run, we collect 5 initial rollouts from the respective environment by randomly sampling from its action space. After every 100th training step, we extend the training set by collecting an additional rollout using the planner, but add a Gaussian distributed exploration noise $\epsilon \sim \mathcal{N}(0, \Unit \cdot 0.3^2)$ to each action. After every 1000th training step, we additionally collect a test rollout for evaluation (which is not added to the training set) using the planner without exploration noise, yielding the results in Fig.~\ref{fig:planet}. Results for additional environments are depicted in Fig.~\ref{fig:planet_additional}. For both train and test collections the planner is identical within each experiment, being either "CEM extensive", "CEM low budget", or "iCEM low budget" (see table~\ref{tab:planet:planner} for details). For each experiment configuration we report results on 3 independent training runs. After training each model for 100k steps, we collect 10 evaluation rollouts (without exploration noise) per training run (i.e., 30 in total) and report the results in table~\ref{tab:performance}.

\begin{table}[htbp]
  \caption{PlaNet CEM details}
  \small
  \begin{center}
    \makebox[0pt]{\begin{tabular}{@{}L@{}ccccccccc@{}}
      \toprule
                & CEM-       & initial    & elites & clip   & best  & decay & reuse   & shift & shift \\
                & iterations & candidates &        & action & action &       & elites & means & elites \\
      \midrule
      CEM extensive   & 10   & 1000 & 100 &  yes  & yes  & 1.0  & no & no & no\\
      CEM low budget  & 3    & 122  & 10  &  yes  & yes  & 1.0  & no & no & no\\
      iCEM low budget & 3    & 150  & 10  &  yes & yes & 1.25 & yes & yes & yes\\
      \midrule
                      &budget & mean as & \multicolumn{2}{c}{momentum}    & \multicolumn{2}{c}{initial std.} & \multicolumn{3}{c}{colored-noise exponent}\\
                      &       & sample  & \multicolumn{2}{c}{$\alpha$}                     & \multicolumn{2}{c}{$\sigma_{init}$}          & \multicolumn{3}{c}{$\beta$} \\
      \midrule
      CEM extensive   &10000 & no   & \multicolumn{2}{c}{0}   & \multicolumn{2}{c}{1.0} & \multicolumn{3}{c}{0}\\
      CEM low budget  &366   & no   & \multicolumn{2}{c}{0}   & \multicolumn{2}{c}{1.0} & \multicolumn{3}{c}{0}\\
      iCEM low budget &366   & yes  & \multicolumn{2}{c}{0.1} & \multicolumn{2}{c}{see table~\ref{tab:hyperparams:env}} & \multicolumn{3}{c}{see table~\ref{tab:hyperparams:env}} \\

        \bottomrule
    \end{tabular}}
  \end{center}
  \label{tab:planet:planner}
\end{table}

\section{Ablation results}\label{sec:sup:ablations}

\begin{figure}[htbp]
  \centering

  \begin{minipage}{.33\linewidth}
    \centering
    Budget 100
  \end{minipage}\hfill
  \begin{minipage}{.33\linewidth}
    \centering
    Budget 300
  \end{minipage}\hfill
  \begin{minipage}{.33\linewidth}
    \centering
    Budget 500
  \end{minipage}\\

  \HCR{}

    \begin{subfigure}{.33\linewidth}
        \centering
        \includegraphics[width=\linewidth]{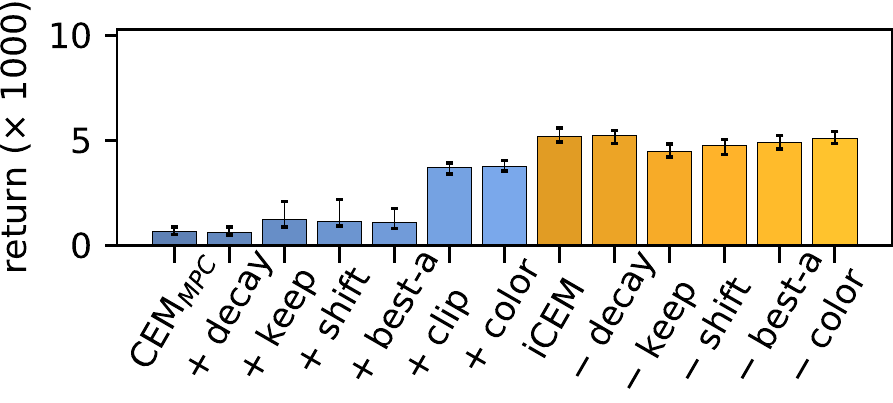}
    \end{subfigure}\hfill
    \begin{subfigure}{.33\linewidth}
        \centering
        \includegraphics[width=\linewidth]{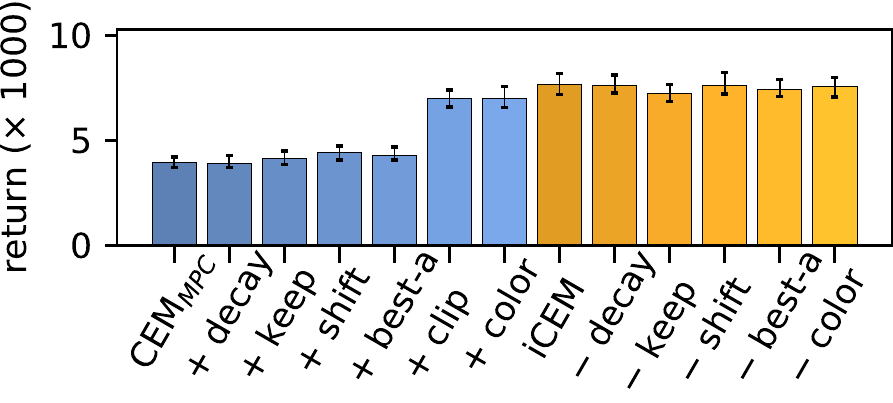}
    \end{subfigure}\hfill
    \begin{subfigure}{.33\linewidth}
        \centering
        \includegraphics[width=\linewidth]{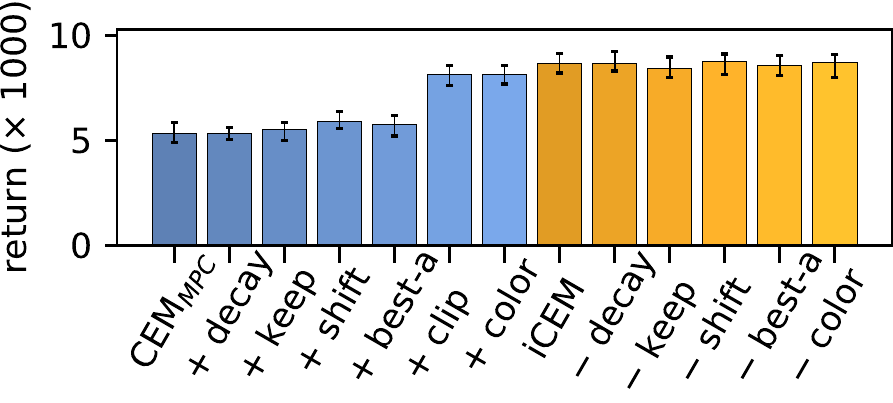}
    \end{subfigure}

    \HS{}

    \begin{subfigure}{.33\linewidth}
        \centering
        \includegraphics[width=\linewidth]{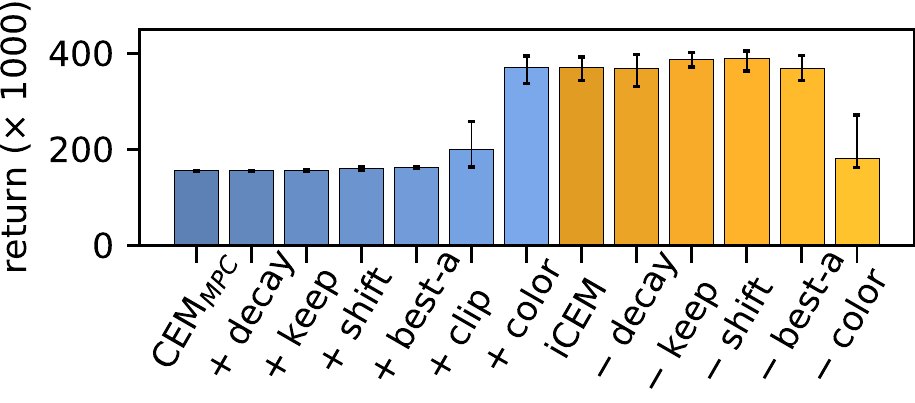}
    \end{subfigure}\hfill
    \begin{subfigure}{.33\linewidth}
        \centering
        \includegraphics[width=\linewidth]{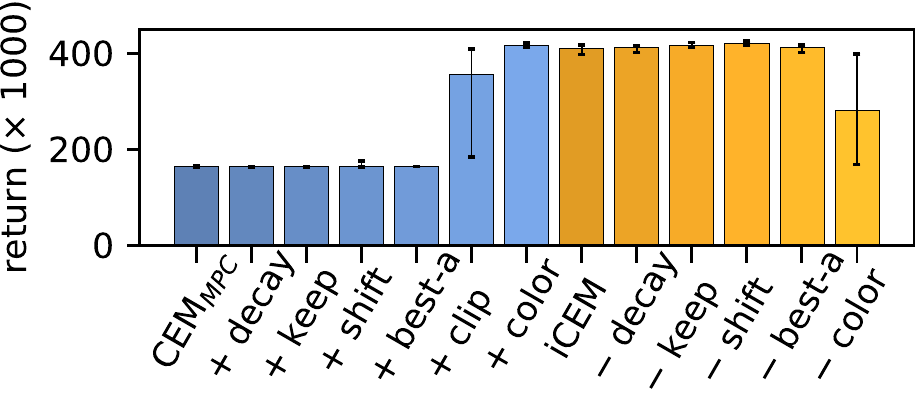}
    \end{subfigure}\hfill
    \begin{subfigure}{.33\linewidth}
        \centering
        \includegraphics[width=\linewidth]{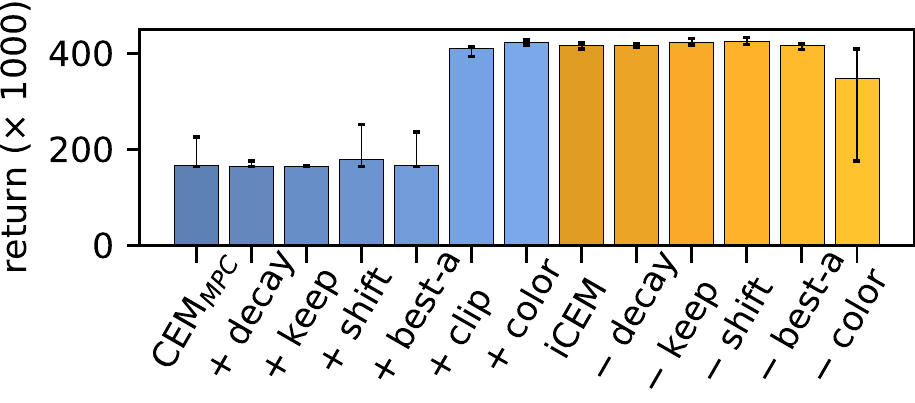}
    \end{subfigure}

    \FPP{}

    \begin{subfigure}{.33\linewidth}
        \centering
        \includegraphics[width=\linewidth]{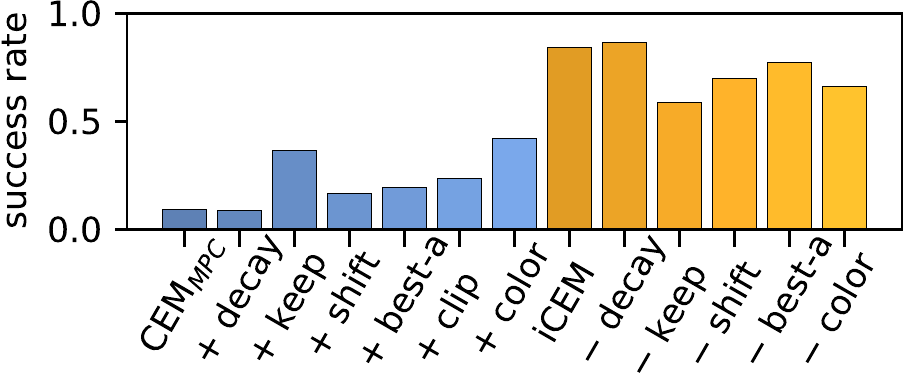}
    \end{subfigure}\hfill
    \begin{subfigure}{.33\linewidth}
        \centering
        \includegraphics[width=\linewidth]{ablation-fpp-ablations_success-300}
    \end{subfigure}\hfill
    \begin{subfigure}{.33\linewidth}
        \centering
        \includegraphics[width=\linewidth]{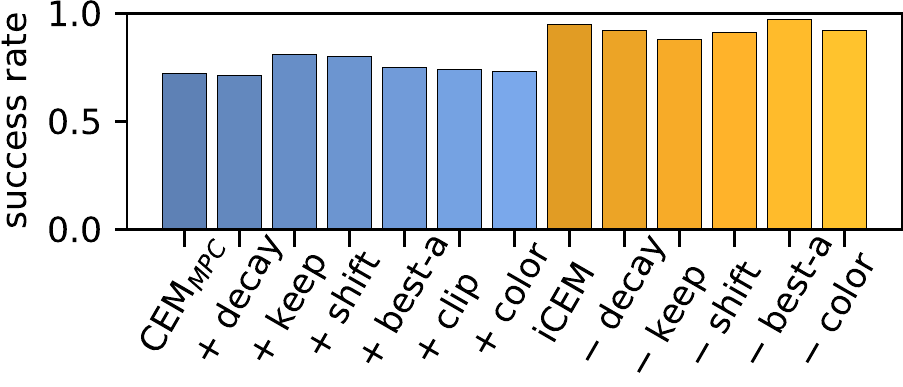}
    \end{subfigure}

    \Door{}

    \begin{subfigure}{.33\linewidth}
        \centering
        \includegraphics[width=\linewidth]{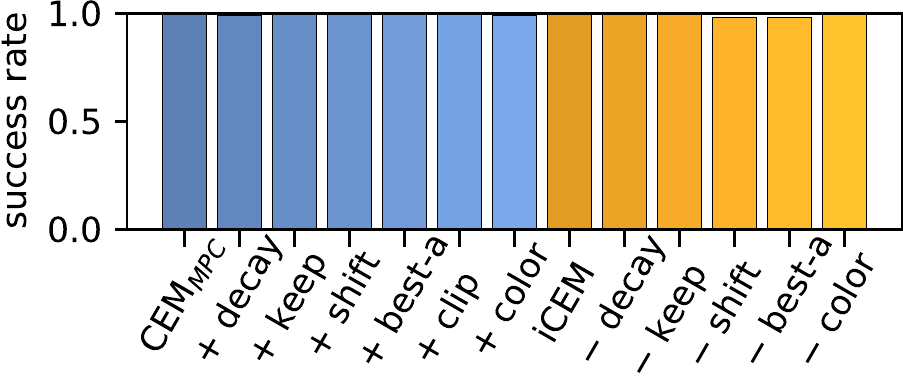}
    \end{subfigure}\hfill
    \begin{subfigure}{.33\linewidth}
        \centering
        \includegraphics[width=\linewidth]{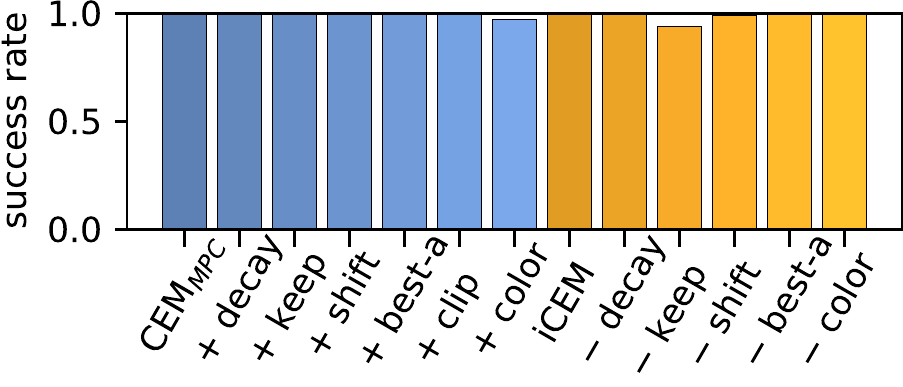}
    \end{subfigure}\hfill
    \begin{subfigure}{.33\linewidth}
        \centering
        \includegraphics[width=\linewidth]{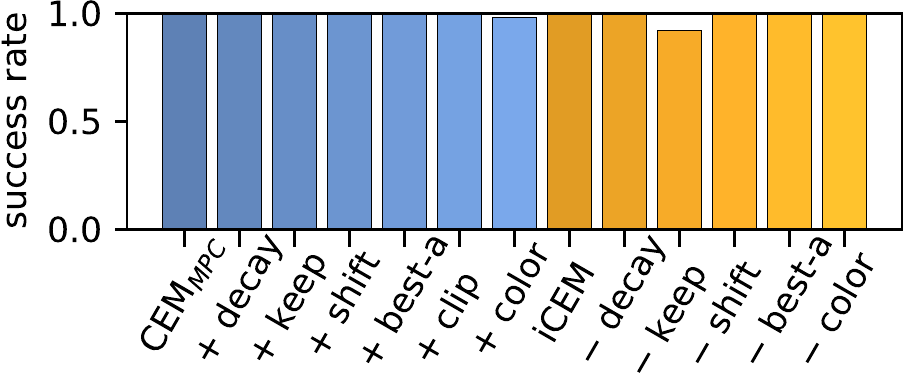}
    \end{subfigure}

    \Door{} (sparse reward)

    \begin{subfigure}{.33\linewidth}
        \centering
        \includegraphics[width=\linewidth]{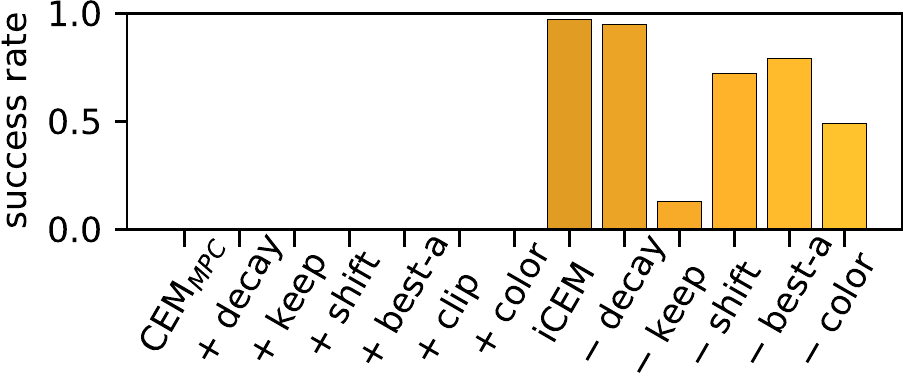}
    \end{subfigure}\hfill
    \begin{subfigure}{.33\linewidth}
        \centering
        \includegraphics[width=\linewidth]{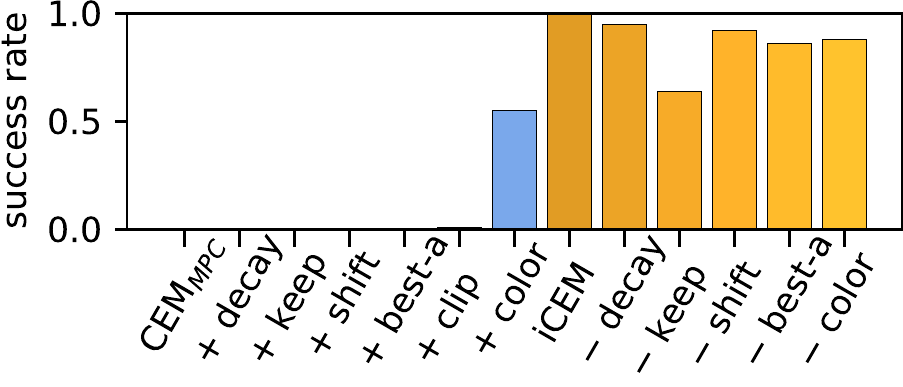}
    \end{subfigure}\hfill
    \begin{subfigure}{.33\linewidth}
        \centering
        \includegraphics[width=\linewidth]{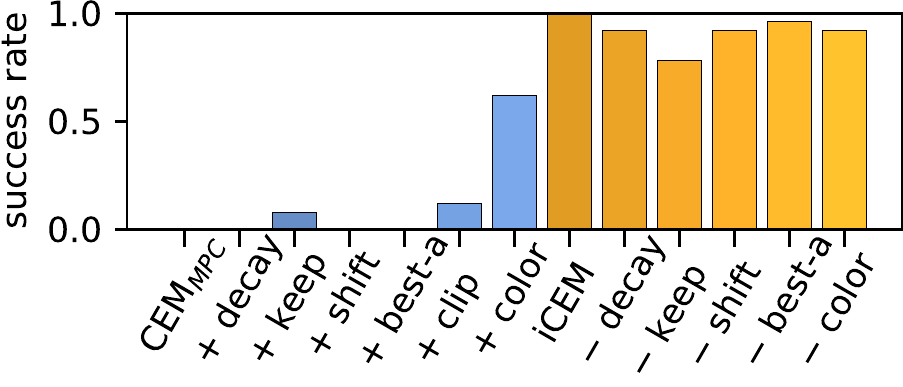}
    \end{subfigure}

    \Reloc{}

    \begin{subfigure}{.33\linewidth}
        \centering
        \includegraphics[width=\linewidth]{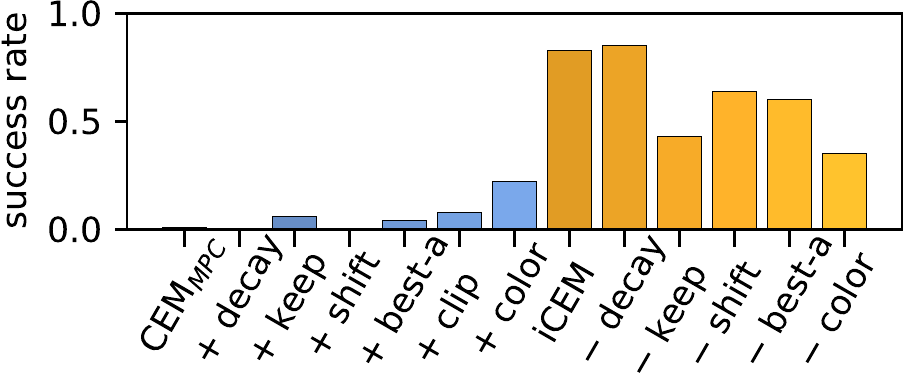}
    \end{subfigure}\hfill
    \begin{subfigure}{.33\linewidth}
        \centering
        \includegraphics[width=\linewidth]{ablation-relocate-ablations-300}
    \end{subfigure}\hfill
    \begin{subfigure}{.33\linewidth}
        \centering
        \includegraphics[width=\linewidth]{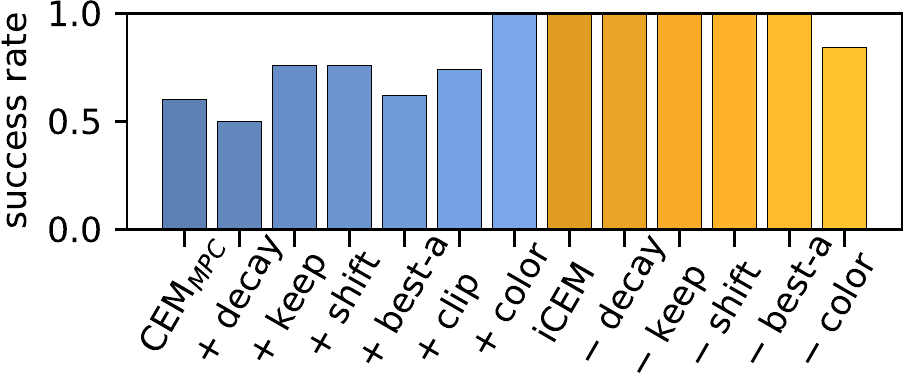}
    \end{subfigure}
    \caption{Ablation studies}
    \label{fig:sup:ablations}
\end{figure}

In \fig{fig:sup:ablations} the ablations and additions are shown for all environments and a selection of budgets. 
As we use the same hyperparameters for all experiments, see \sec{sec:sup:params}, in some environments a few of the ablated versions perform slightly better but overall our final version has the best performance. 
As seen in \fig{fig:sup:ablations}, not all components are equally helpful in the different environments as each environment poses different challenges.
For instance, in \HS{} the optimizer can get easily stuck in a local optimum corresponding to a sitting posture.
Keeping balance in a standing position is also not trivial since small errors can lead to unrecoverable states.
In the \FPP{} environment, on the other hand, the initial exploration is critical since the agent receives a meaningful reward only if it is moving the box. Then colored noise and keep elites and shifting elites is most important.


\section{Details on the \Method{} improvements}\label{sec:sup:improvement-details}
\subsection{Sampling Colored Noise}\label{sec:sup:sample:color}
To sample action sequences with a specific power spectrum we use the efficient implementation of \cite{Timmer1995generatingCN}, which can be found as a python package at \url{https://github.com/felixpatzelt/colorednoise}.

\subsection{Adding the mean actions}\label{sec:sup:add-mean}

As the dimensionality of the action space increases, it gets more and more difficult to sample an action sequence closer to the mean of the distribution.
Nevertheless, executing the mean might be beneficial for many tasks which require ``clean'' action sequences like, for example, manipulation, object-reaching, or any linear trajectory in the state-space.
Adding the mean to the samples fixes this problem and closes the gap with the original CEM, allowing the algorithm to pick either the mean or the best sampled action.

However, we noticed an unexpected performance degradation when adding the mean in every CEM-iteration,
presumably due to the effect of quicker narrowing down the variance along CEM-iterations.
Adding the mean just at the last iteration prevents this bias and has advantageous effects.
If the mean survives the last iteration and becomes part of the elite-set, it will be automatically shifted to the successive time step.

\section{Spectral characteristics of noise}\label{sec:sup:colored-noise}
We can achieve more efficient exploration by choosing different kinds of action noise, which in turn affects the type of correlations between actions at different time steps.
We can notice this by writing down the auto-correlation function which, according to the Wiener-Khinchin theorem, can be expressed as the inverse Fourier transform of the power spectral density of the control input: $C(\tau) = \mathcal{F}^{-1}[\textrm{PSD}_a(f)]$.
If the power spectral density follows the inverse power law of Eq.~\eqref{eq:PSD-noise}, and we apply a scale transformation in the time domain $\tau \rightarrow \tau'= s\tau $, then, from the frequency scaling property of the Fourier transforms:
\begin{flalign*}
&& C(s\tau)&=\mathcal{F}^{-1}\left[\frac{1}{s}\textrm{PSD}_a\left(\frac{f}{s}\right)\right]&\\
&& &=\mathcal{F}^{-1}\left[\frac{1}{s}s^\beta\textrm{PSD}_a(f)\right] \qquad  \text{using Eq.~\eqref{eq:PSD-noise}}\\
&& &=s^{\beta-1}\mathcal{F}^{-1}[\textrm{PSD}_a(f)]\\
&& &=s^{\beta-1} C(\tau)
\end{flalign*}
From this self-referential formula we can understand to which degree the actions lose similarity with a copy of themselves at a different point in time, as detailed in \cite{fossion2010ACFscaling}.

\end{document}